\documentclass{article}

\usepackage{microtype}
\usepackage{graphicx}
\usepackage{booktabs} % for professional tables

\usepackage[hyphens]{url}
\usepackage{hyperref}
\usepackage[accepted]{mlsys2022}

\usepackage{xr}
\usepackage{adjustbox}
\usepackage{amsmath}
\usepackage{wrapfig}
\usepackage{multirow}
\usepackage{algorithm}
\usepackage{algpseudocode}
\usepackage{enumitem}
\setitemize{itemsep=-.5pt,topsep=-.5pt,parsep=0pt,partopsep=0pt}
\usepackage{array}
\usepackage{tikz}
\usepackage{caption}
\usepackage{float}
\usepackage{subcaption}
\usepackage{pifont}
\usepackage{xcolor}
\usepackage{array}
\usepackage{fancyvrb}
\usepackage{fixltx2e}
\usepackage[flushleft]{threeparttable}
\newcolumntype{L}[1]{>{\raggedright\let\newline\\\arraybackslash\hspace{0pt}}m{#1}}
\newcolumntype{C}[1]{>{\centering\let\newline\\\arraybackslash\hspace{0pt}}m{#1}}
\newcolumntype{R}[1]{>{\raggedleft\let\newline\\\arraybackslash\hspace{0pt}}m{#1}}

\newcommand{\Sys}{\textsc{CoRa}}

\newcolumntype{x}[1]{>{\centering\arraybackslash\hspace{0pt}}p{#1}}

\def\zz#1{%
 \ifx\zz#1\else
   #1\linebreak[1]\expandafter\zz
 \fi}

\CustomVerbatimCommand{\ppfverb}{Verb}{commandchars=\\\{\},codes={\catcode`$=3\catcode`_=8}}

\newcommand*\circled[1]{\tikz[baseline=(char.base)]{
            \node[shape=circle,draw,thick,inner sep=1pt, scale = 0.83] (char) {#1};}}
\newcolumntype{L}[1]{>{\raggedright\let\newline\\\arraybackslash\hspace{0pt}}m{#1}}
\newcolumntype{C}[1]{>{\centering\let\newline\\\arraybackslash\hspace{0pt}}m{#1}}

\newif\ifminted
\mintedtrue

\makeatletter\let\expandableinput\@@input\makeatother

\usepackage{listings}

\usepackage{titlesec}
\usepackage{microtype}
\usepackage{relsize}
\usepackage{etoolbox}

\mlsystitlerunning{The \Sys~Tensor Compiler}

\vfuzz \hfuzz
\usepackage{autobreak}
\titlespacing\section{0pt}{12pt plus 4pt minus 2pt}{0pt plus 2pt minus 2pt}
\titlespacing\subsection{0pt}{12pt plus 4pt minus 2pt}{0pt plus 2pt minus 2pt}
\titlespacing\subsubsection{0pt}{12pt plus 4pt minus 2pt}{0pt plus 2pt minus 2pt}
\begin{document}

\setlength{\abovedisplayskip}{3pt}
\setlength{\belowdisplayskip}{3pt}

\twocolumn[
\mlsystitle{The \Sys~Tensor Compiler: \\Compilation for Ragged Tensors with Minimal Padding}

% unless you've provided the [accepted] option to the mlsys2022


\mlsyssetsymbol{equal}{*}

\begin{mlsysauthorlist}
\mlsysauthor{Pratik Fegade}{cmu}
\mlsysauthor{Tianqi Chen}{cmu,octoml}
\mlsysauthor{Phillip B. Gibbons}{cmu}
\mlsysauthor{Todd C. Mowry}{cmu}
\end{mlsysauthorlist}

\mlsysaffiliation{cmu}{Carnegie Mellon University, Pittsburgh, USA}
\mlsysaffiliation{octoml}{OctoML}

\mlsyscorrespondingauthor{Pratik Fegade}{ppf@cs.cmu.edu}

\mlsyskeywords{Machine Learning, MLSys}

\vskip 0.3in

\begin{abstract}
  There is often variation in the shape and size of input data used for
deep learning. In many cases, such data can be represented using
tensors with non-uniform shapes, or \emph{ragged tensors}. Due to
limited and non-portable support for efficient execution on ragged
tensors, current deep learning frameworks generally use techniques
such as padding and masking to make the data shapes uniform and then
offload the computations to optimized kernels for dense tensor
algebra. Such techniques can, however, lead to a lot of wasted
computation and therefore, a loss in performance. This paper
presents \Sys, a tensor compiler that allows users to easily generate
efficient code for ragged tensor operators targeting a wide range of
CPUs and GPUs. Evaluating \Sys~on a variety of operators on ragged
tensors as well as on an encoder layer of the transformer model, we
find that \Sys~(i) performs competitively with hand-optimized
implementations of the operators and the transformer encoder and (ii)
achieves a $1.6\times$ geomean speedup for the encoder on an Nvidia
GPU over PyTorch and a $1.37\times$ geomean speedup for the multi-head
attention module used in transformers on a 64-core ARM CPU over
TensorFlow.

\end{abstract}
]



\printAffiliationsAndNotice{}  % leave blank if no need to mention equal contribution
%% \printAffiliationsAndNotice{\mlsysEqualContribution} % otherwise use the standard text.

\section{Introduction} \label{sec:intro}
Deep learning (DL) is used for a variety of computational tasks on
different kinds of data including sequential data like
text~\cite{treelstm, transformer}, audio~\cite{wavenet} and
music~\cite{music_dl, music_trans} and spatial data like
images~\cite{resnet}. Simultaneously, DL models have become more and
more computationally expensive. More efficient execution of these
models is, therefore, a priority.

\begin{figure}
  \centering
  \includegraphics[width=0.85\columnwidth]{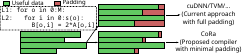}
  \caption{An example operation on ragged tensors.}
  \label{fig:ragged_softmax}
  \vspace{-2mm}
\end{figure}

There is often variation in the sizes of the data that we process
using DL. Images can be of different resolutions, textual sentences
and documents can be of different lengths, and audio can be of
different durations. Processing such data exhibiting variation in
shape, or \emph{shape dynamism}~\cite{nimble}, using the same model
and further, as part of the same mini-batch is therefore important. An
example elementwise operation on such data is shown in
Fig.~\ref{fig:ragged_softmax}, where the slices of the inner dimension
of tensor \verb+A+ have variable sizes. Such tensors and operators are
referred to as \emph{ragged tensors} and \emph{ragged operators}
respectively. Note how the shape dynamism translates to a variable
bound for loop \verb+L2+ which iterates over the variable-sized tensor
slices.

\begin{figure}
  \centering
  \includegraphics[width=0.9\columnwidth]{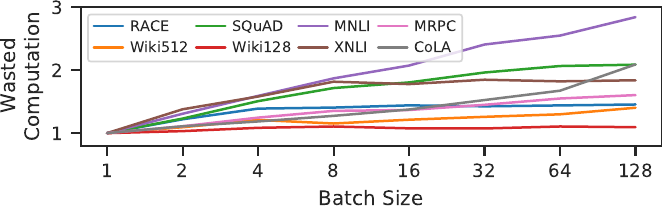}
  \caption{Wasted computation due to padding in a transformer encoder
    layer.}
  \label{fig:flop_ratios}
  \vspace{-2mm}
\end{figure}

%% Padding and masking are commonly used techniques~\cite{pad} to deal
%% with shape dynamism. When data of different structures and/or sizes
%% is batched together in a mini-batch, the individual data points are
%% padded so that all the data points have the same structure as
%% illustrated on the left pane of Fig.~\ref{fig:ragged_softmax}. This
%% makes it possible to then offload the computation to dense tensor
%% algebra kernels either implemented as part of vendor libraries such
%% as Nvidia's cuDNN~\cite{cudnn} and Intel's oneDNN~\cite{onednn} or
%% generated by tensor compilers such as TVM~\cite{tvm} or
%% Halide~\cite{halide}.

Past work has developed hand-optimized kernels to accelerate some
important ragged applications such as batched matrix multiplication
with variable dimensions~\cite{cbt, magma}, triangular matrix
multiplication~\cite{trmm} and the widely-used
transformer~\cite{transformer} models~\cite{EffectiveTrans, FT}. Such
hand-optimized kernels, however, require substantial development
effort and, hence, are available only for a few operators. Further,
they are not portable across different hardware substrates, which is
problematic due to the %rapid pace of innovation in DL hardware.
rapid innovation in DL hardware.

While some DL frameworks have started providing support for ragged
operators recently~\cite{TFRagged, PTNested}, it is quite
limited~\cite{TFIssue, PTIssue, NTL} as discussed
in~\S\ref{sec:related}. Therefore, frameworks usually rely on
efficient dense tensor algebra kernels implemented in vendor libraries
such as cuDNN~\cite{cudnn} and oneDNN~\cite{onednn} or generated by
tensor compilers such as TVM~\cite{tvm} to target parallel
hardware. Padding (illustrated in the top right of
Fig.~\ref{fig:ragged_softmax}) and masking\footnote{Masking involves
setting some tensor elements to a special value so that these elements
are ignored in computations.} are therefore commonly used to eliminate
shape dynamism in ragged tensors and enable the use of vendor
libraries or dense tensor compilers~\cite{HuggingFacePadding}.

Padding and masking, however, lead to wasted computation as the
padding or the masked data points are discarded after execution. %
%% Given the high computational requirements of today's models,
Fig.~\ref{fig:flop_ratios} plots the relative amount of computation
(computed analytically in FLOPs) involved in the forward pass of an
encoder layer of the transformer model\footnote{The hyperparameters
used are the same as those in~\S\ref{sec:layer_eval}.} with and
without padding. We see that padding leads to a significant increase
in the computational requirements of the layer, especially at larger
batch sizes, increasing computation in an already computationally
expensive model.

%% As alluded to before and illustrated in Fig.~\ref{fig:ragged_softmax},
%% a major reason for using padding to handle shape dynamism is because
%% most vendor libraries and tensor compilers only provide support for
%% fully dense linear algebra kernels. %
%% either fully dense linear algebra kernels, or, in some cases, for
%% sparse linear algebra kernels.
%% We note however, that both of these abstractions are a sub-optimal fit
%% for operations on ragged tensors. Ragged tensors are similar to sparse
%% tensors in the sense that iterating over them requires loop nests with
%% variable loop bounds, as Fig.~\ref{fig:ragged_softmax} shows, while
%% they are similar to their fully dense counterparts in the sense that
%% the data is densely packed, as opposed to sparse tensors. Neither of
%% the two extremes of fully dense or fully sparse kernels is therefore a
%% good fit for implementing ragged operations.
Thus, current solutions for efficient ragged operator execution are
unsatisfactory%% , either due to low portability and high development
%% costs, or large amounts of wasteful computation
. Hence, we propose a compiler-based solution enabling easy and more
portable generation of performant code for ragged operators. While
sparse~\cite{comet, taco} and dense~\cite{tvm, tc, halide, tiramisu}
tensor compilers have been well-studied, it is not straightforward to
apply these techniques to ragged tensors, due to the following
challenges:

\begin{enumerate}[label=\textbf{C\arabic*}, leftmargin=1.4em, topsep=0pt, itemsep=-0.25em]
%% \item \label{ch:general} \textbf{Generality of Sparse Tensor Compilers:}
%%   Sparse tensor compilers usually have to support a wide variety of
%%   storage formats that are used to store sparse tensors. This
%%   generality often means that it is difficult to generate highly
%%   specialized code for different storage formats.

\item \label{ch:irregular} \textbf{Irregularity in generated code:}
  While the data in ragged tensors are densely packed, the variable
  loop bounds can lead to irregular code, often causing a loss of
  performance on hardware substrates such as GPUs.

\item \label{ch:abstract} \textbf{Insufficient compiler mechanisms:}
  Representing transformations on loops with variable bounds and on
  tensor dimensions with variable-sized slices is not straightforward
  due to the dependencies that exist among loops and tensor dimensions
  respectively in ragged operators. Further, optimization decisions
  made by sparse tensor compilers may not always work for ragged
  tensors as sparse tensors are much sparser than ragged tensors.

\item \label{ch:general} \textbf{Ill-fitting computation
  abstractions:} There is a mismatch between the interfaces and
  abstractions provided by current compilers and ragged
  operators. Such operators cannot be expressed in dense compilers,
  while sparse compilers do not adequately provide ways to express
  information relevant to efficient code generation.

  %% Computations on ragged tensors differ from the dense and sparse
  %% cases in important ways. Using current compilers for these
  %% computations leads to sub-optimal outcomes such as the use of
  %% padding or the use of sub-optimal storage formats in the case of
  %% sparse tensor compilers.
\end{enumerate}

%% With these challenges in mind, we present
%% \Sys\footnote{\textbf{Co}mpiler for \textbf{Ra}gged Tensors}, a
%% tensor compiler built for ragged tensors. \Sys~allows one to
%% express computations on ragged tensors, schedule them and target a
%% variety of different substrates such as CPUs and GPUs. To overcome
%% challenge~\ref{ch:general}, as part of \Sys, we design an interface
%% for the user to express ragged operations, as well as provide the
%% compiler with the knowledge necessary for generating efficient
%% code, %% statically storage access lowering scheme
%% (\S\ref{sec:lower_storage}) %% specialized for ragged tensors. This
%% allows us to lower tensor %% accesses involving indirect memory
%% accesses in an efficient manner.  This includes the ability to
%% strategically pad dimensions of ragged tensors
%% (\S\ref{sec:scheduling}) and convey the information to the compiler
%% as well as to specify thread remapping strategies to lower load
%% imbalance (\S\ref{sec:scheduling}), thereby also tackling
%% challenge~\ref{ch:irregular}. To overcome~\ref{ch:abstract},
%% \Sys~uses uninterpreted functions for cleanly representing variable
%% loop bounds and scheduling operations on the same
%% (\S\ref{sec:fusion}). This allows us to systematically handle
%% ragged loop nests and ragged storage formats.

\newcommand{\ghigh}[0]{\textcolor{green}{High}}
\newcommand{\rfull}[0]{\textcolor{red}{Full}}
\newcommand{\rlow}[0]{\textcolor{red}{Low}}
\newcommand{\rhigh}[0]{\textcolor{red}{High}}
\newcommand{\glow}[0]{\textcolor{green}{Low}}
\newcommand{\gminimal}[0]{\textcolor{green}{Minimal}}
\begin{table}[t]
\vspace{-2.5mm}
\caption{Comparison between \Sys~and current solutions for ragged
  operations. TC stands for tensor compilers.}
  \vspace{-1.5mm}
\label{table:comp}
\begin{center}
\begin{scriptsize}
\addtolength{\tabcolsep}{-4pt}
\resizebox{0.99\columnwidth}{!}{%
\begin{tabular}{C{2.4cm}|C{1.1cm}C{1.5cm}C{0.9cm}C{1.3cm}}
\toprule
Framework               & Portability            & Operator impl.~effort             & Padding   & Performance \\
\midrule
Dense TC                & \ghigh                 & \glow                             & \rfull    & \rlow       \\
Sparse TC               & \ghigh                 & \glow                             & \gminimal & \rlow       \\
Dense vendor libs.      & \rlow                  & \rhigh                            & \rfull    & \rlow      \\
Hand-optimized impl.    & \rlow                  & \rhigh                            & \gminimal & \ghigh      \\
\midrule
\Sys                    & \ghigh                 & \glow                             & \gminimal & \ghigh      \\
\bottomrule
\end{tabular}
}
\addtolength{\tabcolsep}{4pt}
\end{scriptsize}
\end{center}
\vspace{-1.5mm}
\end{table}

With these challenges in mind, we present \Sys~(\textbf{Co}mpiler for
\textbf{Ra}gged Tensors), a tensor compiler which allows one to
express and optimize ragged operations to easily target a variety of
substrates such as CPUs and GPUs.
%% statically storage access lowering scheme (\S\ref{sec:lower_storage})
%% specialized for ragged tensors. This allows us to lower tensor
%% accesses involving indirect memory accesses in an efficient manner.
To overcome challenge~\ref{ch:irregular}, \Sys~enables minimal padding
of ragged tensor dimensions (\S\ref{sec:scheduling}) in order to
generate efficient code for targets such as GPUs as well as to specify
thread remapping strategies to lower load imbalance
(\S\ref{sec:scheduling}). \Sys~uses uninterpreted
functions~\cite{spf1} to symbolically represent variable loop bounds
and scheduling operations on the same (\S\ref{sec:fusion}). %% This
%% allows us to systematically
%% handle ragged loop nests and ragged storage formats.
\Sys's mechanisms (such as its storage lowering scheme discussed
in~\S\ref{sec:lower_storage}) and optimizations are specialized for
ragged tensors thereby tackling~\ref{ch:abstract}. Further,
\Sys~provides simple abstractions to convey information essential to
efficient code generation such as padding or thread remapping
specifications and raggedness patterns of tensors to the compiler
(\S\ref{sec:api}). This overcomes challenge~\ref{ch:general}.

\Sys~enables efficient code generation for ragged operators by
significantly reducing %% wasted computation due to
padding (\S\ref{sec:eval}). %
%% Further, storing tensors without full padding also leads to lower
%% memory usage (\S\ref{sec:layer_eval}).
As part of \Sys's implementation, we reuse past work by extending a
tensor compiler~\cite{halide, tvm, tiramisu, taco} and thus, provide
familiar interfaces to \Sys's users. This also makes it easy in the
future to use auto-scheduling~\cite{halide_auto1, halide_auto2,
  tvm_autotune, ansor, auto_sched10} for optimizing ragged tensor
operations. Table~\ref{table:comp} compares \Sys~with alternatives
that are or could be used for ragged operators.  Only \Sys{} achieves
high performance and portability, with low operator implementation
effort (and minimal padding).

In summary, this paper makes the following contributions:
\begin{enumerate}[topsep=0pt, itemsep=-0.25em, leftmargin=1.25em]
  \item We present \Sys, a tensor compiler for ragged tensors. %% that
    %% allows one to express, optimize and generate efficient code for
    %% computations on ragged tensors.
    To our knowledge, \Sys~is the first tensor compiler that allows
    efficient computation on ragged tensors.

  \item As part of the design, we generalize the API, abstractions and
    the mechanisms of tensor compilers and propose new scheduling
    primitives for ragged tensors.

  %% \item We prototype and evaluate \Sys~on a variety of compute
  %%   kernels comparing against state-of-the-art vendor libraries and
  %%   hand-optimized implementations for ragged tensors. We obtain
  %%   overall speedups of $1.6\times$ over PyTorch for a transformer
  %%   encoder layer and perform competitively with a hand-optimized
  %%   implementation on a Nvidia V100 GPU. On an ARM CPU, we perform
  %%   $1.86\times$ and $1.89\times$ better than PyTorch and
  %%   TensorFlow respectively on the masked multi-head attention
  %%   (MHA)~\cite{transformer} module used in transformers.

  %% \item We evaluate \Sys~on a variety of ragged operators. For a
  %%   transformer encoder layer, we perform $1.6\times$ better than
  %%   PyTorch~\cite{PyTorch} and as well as FasterTransformer~\cite{FT},
  %%   a highly optimized transformer implementation, on an Nvidia V100
  %%   GPU. On a 64-core ARM CPU, we obtain a geomean speedup of $1.36$
  %%   over TensorFlow~\cite{TensorFlow}, on the multi-head attention
  %%   (MHA) module~\cite{transformer} used in transformers.

  \item We evaluate \Sys~on a variety of ragged operators. For a
    transformer encoder layer, we perform $1.6\times$ better than
    PyTorch~\cite{PyTorch} and as well as FasterTransformer~\cite{FT},
    a highly optimized transformer implementation, on an Nvidia V100
    GPU. On a 64-core ARM CPU, we are $1.37\times$ faster than
    TensorFlow~\cite{TensorFlow}, for the multi-head attention (MHA)
    module~\cite{transformer} used in transformers.
\end{enumerate}

\section{\Sys{} Overview} \label{sec:overview}
\begin{figure}
  \centering
  \includegraphics[width=0.99\columnwidth]{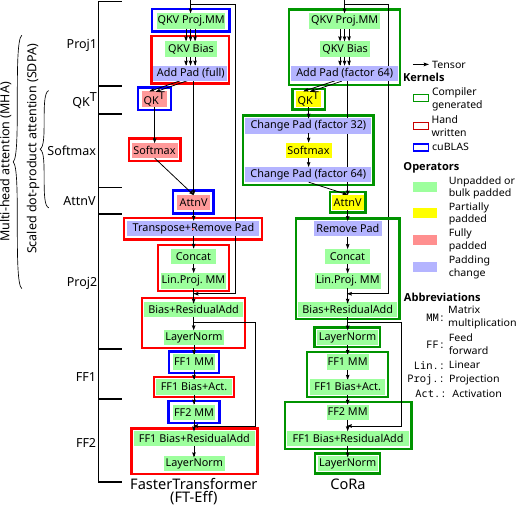}
  %% \caption{FasterTransformer (FT-Eff) and \Sys~implementations of a
  %%   transformer's encoder layer. Note how \Sys's~compiler-based
  %%   implementation uses only partial padding for SDPA as opposed to
  %%   FasterTransformer's fully padded implementation. Note also how
  %%   \Sys~allows one to exploit more operator fusion opportunities
  %%   (including fusing all the layout change operations that add or
  %%   remove padding) as opposed to FasterTransformer, which cannot do
  %%   so in all cases due to its reliance on vendor libraries.}
  \caption{FasterTransformer (FT-Eff) and \Sys~implementations of a
    transformer's encoder layer. Note how \Sys's~fully compiler-based
    implementation uses only partial padding for SDPA as opposed to
    FasterTransformer's fully padded implementation. \Sys~also enables
    more operator fusion (including fusing all the padding change
    operations) as opposed to FasterTransformer, which cannot do so in
    all cases as it relies on vendor libraries.}
  \label{fig:fusion_graph}
  \vspace{-2mm}
\end{figure}

\Sys's compiler-based approach enables the generation of performant
code in a portable manner. This is reflected in
Fig.~\ref{fig:fusion_graph}, which compares \Sys's implementation of a
transformer encoder layer with FasterTransformer.
%% Nvidia's manually-optimized transformer implementation.
The highly-optimized FasterTransformer relies heavily on kernels
implemented in cuBLAS (Nvidia's BLAS library), which are shown as blue
outlines in the figure, and on manually implemented kernels, shown as
red outlines. On the other hand, \Sys's implementation exclusively
employs compiler generated kernels (shown as green outlines), making
it more portable. Further, \Sys's compiler approach allows it to
exploit more kernel fusion opportunities, evident from the fact that
\Sys's implementation launches nine kernels as opposed to
FasterTransformer's twelve. Both the implementations in the figure use
minimal padding for all operators except for those in the scaled
dot-product attention (SDPA) sub-module, where \Sys's specialized
approach enables it to get away with lower padding as compared to
FasterTransformer. We further discuss these implementations
in~\S\ref{sec:eval}.
%% \ppf{Discuss implementation effort?}

%% Shape dynamism in computation involving ragged tensors manifests in
%% two ways---the loop nest can contain loops the bounds of which
%% depend on the iteration variables of outer loops, and the tensors
%% themselves can be stored in a ragged fashion in memory. We saw
%% these features illustrated in Fig.~\ref{fig:ragged_softmax}. When
%% generating code for such computations, \Sys~exploits the following
%% insights:

\Sys's ability to generate performant code that employs minimal
padding in a portable manner as we saw above relies on the following
two insights:
\begin{enumerate}[label=\textbf{I\arabic*}, leftmargin=1.5em, topsep=-1pt, itemsep=-0.25em]
%% \item \label{prop:known_raggedness} Irregularity in ragged tensors
%%   usually arises from the structure of the underlying data. Therefore,
%%   it is usually the case that the pattern of raggedness is known
%%   before the tensor is actually materialized as part of the
%%   computation. The pattern of raggedness is also often the same across
%%   the multiple tensors involved in an operation for the same reason.
\item \label{prop:known_raggedness} In ragged operations, the pattern
  of raggedness is usually known before the tensor is actually
  computed, and is the same across multiple tensors involved in the
  operation.
\item \label{prop:o1_access} Ragged tensors, like dense tensors, allow
  $O(1)$ accesses (\S\ref{sec:lower_storage}). This is unlike sparse
  formats such as compressed sparse row (CSR), where accesses require
  a search over an array. The HASH~\cite{taco_format} sparse format,
  while allowing $O(1)$ accesses, is unsuitable for accelerators such
  as GPUs due to its highly irregular storage.
\end{enumerate}

Insight~\ref{prop:known_raggedness} allows \Sys~to precompute the
auxiliary data structures needed to access ragged tensors without
knowledge of the computation (or values of its input tensors) that
produces the ragged tensor. This and insight~\ref{prop:o1_access}
enable \Sys~to generate efficient code for ragged operations.
%% Further, as we will see in~\S\ref{sec:api}, \Sys~allows user to
%% schedule ragged loop nests by generalizing the scheduling primitives
%% used in tensor compilers for loop nests with constant loop bounds.

\begin{figure*}
  \vspace{-1mm}
  \centering \includegraphics[width=0.9\linewidth]{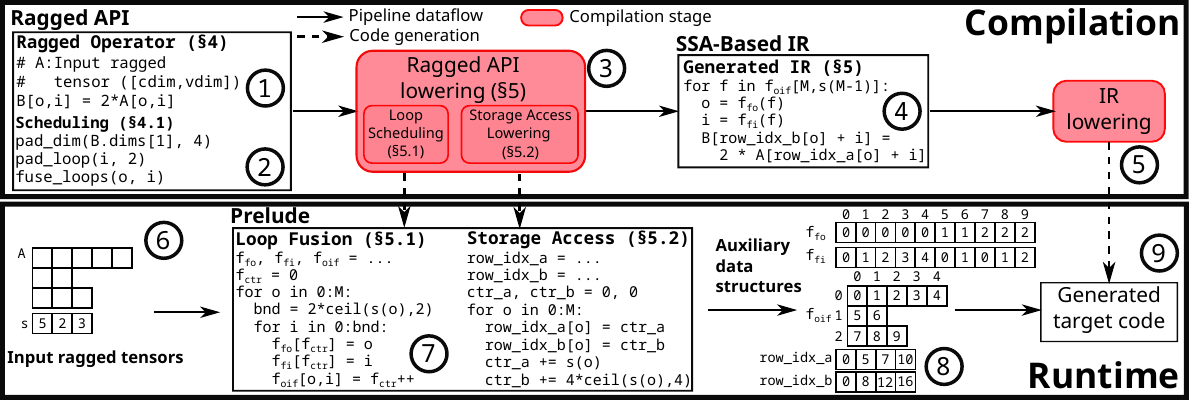}
  \caption{Overview of \Sys's compilation and runtime pipeline.}
  \label{fig:overview}
\end{figure*}

Let us now look at \Sys's overall compilation and execution pipeline,
as illustrated in Fig.~\ref{fig:overview}. The user first expresses
\circled{1} and schedules \circled{2} their computation using an API
similar to that of past tensor compilers (\S\ref{sec:api}). %
%% The user can also specify how the tensors involved are stored in
%% memory.
This specification of the computation and the scheduling primitives
are then lowered \circled{3} to an SSA-based IR \circled{4}. As part
of this lowering step, \Sys~generates code \circled{7} to initialize
some auxiliary data structures it needs to be able to lower accesses
to ragged tensors (\S\ref{sec:lower_storage}) and to enable loop
fusion in ragged loop nests (\S\ref{sec:fusion}). We refer to this
code as the \emph{prelude} code. Compilation then continues with
\Sys~lowering tensor accesses to raw memory offsets by making use of
the data structures generated by the prelude. Finally, \Sys~generates
\circled{5} target-dependent code \circled{9} such as C or CUDA
C++. During execution, the formats of the input ragged tensors
\circled{6} are first processed by the generated prelude code
\circled{7} which creates the auxiliary data structures
\circled{8}. This prelude code is not computationally expensive
(\S\ref{sec:cora_oh_eval}) and hence is executed on the host
CPU. These data structures and the ragged tensors are then passed to
the generated target dependent code \circled{9} which executes on
devices such as CPUs or GPUs.

We will now look these stages in more detail below.

\section{Terminology} \label{sec:term}
Ragged operators have one or more loops with bounds that are functions
of iteration variables of outer loops. We refer to such loops as
\emph{variable loops} or \emph{vloops} while loops with constant
bounds are referred to as constant loops, or \emph{cloops}. A loop
nest with at least one vloop is referred to as a vloop nest%% and a
%% loop nest comprising only of cloops is referred to as a cloop nest
. Further, tensors can be stored in memory with or without
padding. When stored without full padding, the size of some tensor
dimensions depends on outer tensor dimensions. Such dimensions are
referred to as \emph{variable dimensions}, or \emph{vdims} and those
with constant sizes are \emph{constant dimensions} or \emph{cdims}. A
tensor stored such that it has no vdim (i.e. a fully padded tensor) is
referred to as a dense tensor, while a tensor with at least one vdim
is a ragged tensor. Note that ragged tensors may be padded to some
extent.

\section{\Sys's Ragged API} \label{sec:api}
\Sys~provides a simple API similar to that of past tensor compilers,
as seen in Listing~\ref{code:ragged_softmax}, which expresses the
example computation from Fig.~\ref{fig:ragged_softmax} in \Sys. Apart
from describing the computation as in a dense tensor compiler,
\Sys~also requires the user to specify the raggedness dependences of
the computation (highlighted in
Listing~\ref{code:ragged_softmax}). This involves specifying vloop
bounds as functions of outer loop variables and vdim extents as
functions of indices of outer tensor dimensions. Given this
information, \Sys~automatically computes any derived data structures
required (\S\ref{sec:lowering}), making it easy for users to express
their computations. \Sys~uses identifiers called \emph{named
dimensions} (discussed further in~\S\ref{sec:bounds_inference}) to
name loops and corresponding tensor dimensions and to specify
relationships between them. For example, the loop extent defined on
line%
\ifminted
~\ref{line:lext}
\else
~7
\fi%
in the listing states the dependence on the outer
loop, referred to by the named dimension \verb+batch_dim+.

%% %% \ifminted
%% \definecolor{lightyellow}{RGB}{255,255,150}
%% \begin{listing}[h]
%%   \begin{minted}[escapeinside=||,
%%       linenos,
%%       numbersep=4pt,
%%       xleftmargin=\parindent,
%%       frame=lines,
%%       highlightlines={4,6-8,10-11,13,18-19},
%%       highlightcolor=lightyellow,
%%       fontsize=\scriptsize]{src/ra_lexer.py:RALexer -x}
%% ################ Operator Description ################
%% batch_size = var('M')
%% # Declare named dimensions
%% batch_dim, len_dim = Dim(), Dim()
%% # Loop: Specify vloop extents
%% lens = input_tensor((batch_size,))
%% l_ext = Extent([batch_dim], lambda b: lens[b]) |\label{line:lext}|
%% loop_exts = [batch_size, l_ext]
%% # Storage: Specify vdim extents
%% s_ext = Extent([batch_dim], lambda b: lens[b]) |\label{line:sext}|
%% storage_format = [batch_size, s_ext]
%% # Define input ragged tensor
%% dims = [batch_dim, len_dim]
%% A = input_tensor(dims, storage_format)
%% # Express computation
%% B = compute(dims, loop_exts, lambda i,j: 2*A[i,j])
%% ############### Scheduling primitives ###############
%% pad_loop(B.loops[1], 2)                        |\label{line:lpad}|
%% pad_dimension(B.dimensions[1], 4)              |\label{line:spad}|
%% fuse_loops(B.loops[0], B.loops[1])
%%   \end{minted}
%%   \vspace{-3mm}
%%   \captionof{listing}{Operator in Fig.~\ref{fig:ragged_softmax}
%%     expressed in a simplified version of \Sys's API.}
%%   \vspace{-2mm}
%% \label{code:ragged_softmax}
%% \end{listing}

%% \else

\lstset
{ %Formatting for code in appendix
    language=Python,
    basicstyle=\footnotesize,
    numbers=left,
    stepnumber=1,
    showstringspaces=false,
    tabsize=1,
    breaklines=true,
    breakatwhitespace=false,
}

%% \begin{listing}[h]
%%   \begin{lstlisting}[language=Python,basicstyle=\scriptsize]
%% ################ Operator Description ################
%% BS = var('M')                          # Batch size
%% # Declare named dimensions
%% batch_dim, len_dim = Dim(), Dim()
%% # Loop: Specify vloop extents
%% lens = input_tensor((BS,))
%% l_ext = Extent([batch_dim], lambda b: lens[b])
%% loop_exts = [BS, l_ext]
%% # Storage: Specify vdim extents
%% s_ext = Extent([batch_dim], lambda b: lens[b])
%% storage_format = [BS, s_ext]
%% # Define input ragged tensor
%% dims = [batch_dim, len_dim]
%% A = input_tensor(dims, storage_format)
%% # Express computation
%% O = compute(dims, loop_exts, lambda i,j: 2*A[i,j])
%% ############### Scheduling primitives ###############
%% pad_loop(O.loops[1], 32)
%% pad_dimension(O.dimensions[1], 64)
%% fuse_loops(O.loops[0], O.loops[1])
%% \end{lstlisting}
%% \captionof{listing}{Operator in Fig.~\ref{fig:ragged_softmax}
%%   expressed in a simplified version of \Sys's API.}
%% \label{code:ragged_softmax}
%% \end{listing}

  \begin{lstlisting}[language=Python,basicstyle=\scriptsize, captionpos=b, caption={Operator in Fig.~\ref{fig:ragged_softmax} expressed in a simplified version of \Sys's API.}, label={code:ragged_softmax}]
################ Operator Description ################
batch_size = var('M')
# Declare named dimensions
batch_dim, len_dim = Dim(), Dim()
# Loop: Specify vloop extents
lens = input_tensor((batch_size,))
l_ext = Extent([batch_dim], lambda b: lens[b])
loop_exts = [batch_size, l_ext]
# Storage: Specify vdim extents
s_ext = Extent([batch_dim], lambda b: lens[b])
storage_format = [batch_size, s_ext]
# Define input ragged tensor
dims = [batch_dim, len_dim]
A = input_tensor(dims, storage_format)
# Express computation
B = compute(dims, loop_exts, lambda i,j: 2*A[i,j])
############### Scheduling primitives ###############
pad_loop(B.loops[1], 2)
pad_dimension(B.dimensions[1], 4)
fuse_loops(B.loops[0], B.loops[1])
\end{lstlisting}

%% \fi

\subsection{Scheduling Primitives} \label{sec:scheduling}
In order to optimize the expressed computation, \Sys~provides all the
scheduling primitives commonly found in tensor compilers. Below, we
describe some salient features and points of departure from past
tensor compilers.

\noindent\textbf{Loop Scheduling:} Both cloops and vloops can be
scheduled in \Sys. We saw how a vloop, say $L_v$, has a loop bound that
is a function of the iteration variables of one or more outer loops,
say $L_1$ to $L_k$. \Sys~currently does not allow reordering such a
loop $L_v$ beyond any of the loops $L_1$ to $L_k$. %
%% Such a reordering would introduce invalid iterations in the iteration
%% space that would need to be skipped over with the used of conditional
%% expressions.
While possible with the introduction of conditional statements, we
have not found a use case for such reordering.

\begin{figure}
  \centering
  \includegraphics[width=0.9\columnwidth]{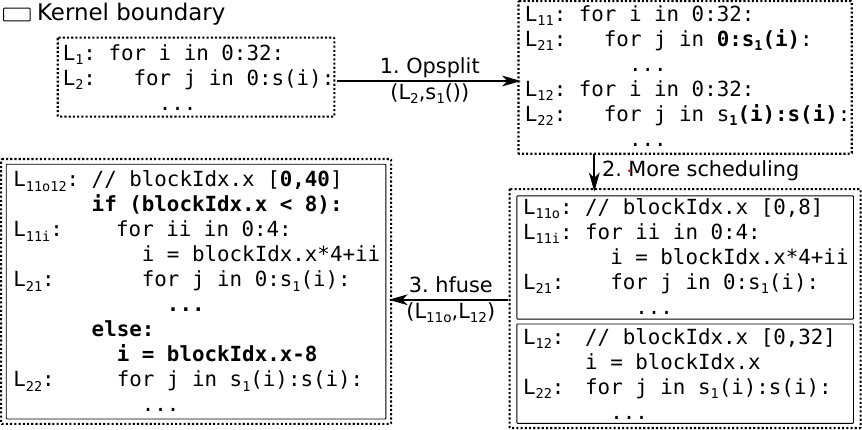}
  \caption{Operation splitting and horizontal fusion. Loop
    \texttt{L}$_{\mathtt{2}}$ is first split in step 1 using operation
    splitting thus creating two loop nests, which are then
    horizontally fused together (step 3) so they execute concurrently
    as part of single kernel.}
  \label{fig:split_hfusion}
\end{figure}

\noindent\textbf{Operation Splitting:} It can sometimes be beneficial
to differently schedule different iterations of a loop in a vloop nest
in order to more optimally handle the variation in loop bounds. %
%% For example, past work~\cite{cbt} on optimizing batched
%% GEMM kernels where each GEMM operation might have different dimensions
%% has effectively used different tile sizes for GEMMs of different
%% dimensions to obtain better performance. To enable such
%% transformations,
\Sys~allows one to split an operation into two or more operations by
specifying split points for one or more of its loops, as
Fig.~\ref{fig:split_hfusion} shows. In our evaluation
(\S\ref{sec:bp_eval}), we use this transformation in conjunction with
horizontal fusion (described below) to better handle the last few
iterations of a tiled loop without the need for additional padding in
the QK\textsuperscript{T} and AttnV operators in the transformer layer
(Fig.~\ref{fig:fusion_graph}).

\noindent\textbf{Horizontal Fusion:} Past work~\cite{hfusion} has
proposed horizontal fusion, or \emph{hfusion} for short, as an
optimization to better utilize massively parallel hardware devices
such as GPUs by executing multiple operators concurrently as part of a
single kernel. With \Sys, we implement this optimization in a tensor
compiler for the outermost loop of two or more operators. HFusion
enables the concurrent execution of the multiple operators that result
from using the operation splitting transform described above.

\noindent\textbf{Loop and Storage Padding:} Despite the overheads of
padding, a small amount of it is often useful in order to generate
efficient vectorized and tiled code by eliding conditional
checks. Accordingly, \Sys~allows the user to specify padding for
vloops and vdims as multiples of a constant. For example, on
line%
\ifminted
~\ref{line:lpad}
\else
~18
\fi%
of Listing~\ref{code:ragged_softmax}, the vloop associated with the
dimension \verb+len_dim+ is asked to be padded to a multiple of 2
while the corresponding dimension of the output tensor is specified to
be padded to a multiple of 4 on line%
\ifminted
~\ref{line:spad}.
\else
~19.
\fi%
Such independent padding specification for loops
and the underlying storage is allowed as long as the storage padding
is at least as much as the loop padding (this ensures that the padded
loop nest never accesses non-existent storage). This ability
allows~\Sys~to fuse padding change operators as is illustrated in
Fig.~\ref{fig:fusion_graph}. We show in~\S\ref{sec:cora_oh_eval} that
this partial padding does not lead to much wasted computation.

\noindent\textbf{Tensor Dimension Scheduling:} \Sys~allows users to
split, fuse and reorder dimensions of dense and ragged tensors. This
can enable more optimal memory accesses. Fusing tensor dimensions in a
way that mirrors the surrounding loop nest can allow for simpler
memory accesses (\S\ref{sec:fusion}).

\noindent\textbf{Load Balancing:} The variable loop bounds in a vloop
nest can lead to unbalanced load across execution units. As proposed
by past work~\cite{dl_sparse} on sparse tensor algebra, \Sys~allows
the user to redistribute work across different parallel processing
elements by specifying a \textit{thread remapping policy}. Given a parallel
loop, this allows the user to specify a mapping between the loop
iterations and the thread id (illustrated in
Fig.~\ref{fig:ap_thread_remap} in the appendix). Depending on the
hardware thread scheduling policy, this can influence the order in
which iterations of the loop are scheduled and lead to non-trivial
performance gains as shown in~\S\ref{sec:vtrmm_eval}.

In conclusion, \Sys~provides familiar and simple interfaces to users,
extended with a few abstractions and scheduling primitives specific to
ragged tensors, enabling their application to support (efficient)
ragged operations.

\section{\Sys's Ragged API Lowering} \label{sec:lowering}
We now discuss some aspects of \Sys's Ragged API lowering that generates
the SSA-based IR as shown in Fig~\ref{fig:overview}.

\subsection{Loop and Tensor Dimension Fusion} \label{sec:fusion}
\begin{figure}
  \centering
  \includegraphics[width=0.85\columnwidth]{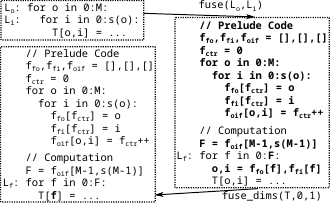}
  \caption{Fusing vloops and tensor dimensions}
  \vspace{-1mm}
  \label{fig:fusion}
\end{figure}

Consider the ragged loop nest shown on the top left corner of
Fig~\ref{fig:fusion}. The loop bound of the inner loop
\texttt{L\textsubscript{i}} is a function $s()$ of \texttt{o}, the
iteration variable of the outer loop \texttt{L\textsubscript{o}}. The
loop \texttt{L\textsubscript{f}} obtained by fusing
\texttt{L\textsubscript{o}} and \texttt{L\textsubscript{i}} is shown
on the right of the figure. The loop bound \texttt{F} of the fused
loop would be equal to $\sum_{o=0}^{M-1}s(o)$. Further note that while
we have fused the loops \texttt{L\textsubscript{o}} and
\texttt{L\textsubscript{i}}, the tensor access \texttt{T[o,i]} in the
body of the loop nest still uses variables \texttt{o} and
\texttt{i}. Therefore, we need to compute the values of these two
variables corresponding to the current value of \texttt{f}, the
iteration variable of \texttt{L\textsubscript{f}}. Because of the
ragged nature of the loop nest, computing the loop bound \texttt{F} as
well as the mapping between the iteration variables of the original
and the fused loop nests is not straightforward. In~\Sys, we generate
code to compute these quantities and variable relationships (shown in
the right pane of Fig.~\ref{fig:fusion}) as part of the prelude which
executes before the main kernel computation. We use vloop fusion as
described above to implement the linear transformation operators
(Proj1, Proj, FF1 and FF2) in the transformer encoder
(Fig.~\ref{fig:fusion_graph}) with minimal padding.

%% Suppose now that the tensor \verb+T+ in Fig.~\ref{fig:fusion} has a
%% storage format that mirrors the loop nest consisting of the unfused
%% loops \verb+Lo+ and \verb+Li+. This means that the 2-dimensional
%% tensor has an outer cdim and an inner vdim the size of the $i^{th}$
%% slice of which is $s(i)$. Fusing the two dimensions of \verb+T+ would
%% then enable \Sys~to simplify the tensor access as shown in the bottom
%% left pane of the figure.

Suppose now that the tensor \texttt{T} in Fig.~\ref{fig:fusion} has a
storage format that mirrors the loop nest consisting of
\texttt{L\textsubscript{o}} and \texttt{L\textsubscript{i}}. This
means that the 2-dimensional tensor has an outer cdim and an inner
vdim the size of the $i^{th}$ slice of which is $s(i)$. Fusing these
dimensions then enables \Sys~to simplify the tensor access as shown in
the bottom left pane of the figure.

\subsection{Bounds Inference}\label{sec:bounds_inference}

\begin{figure}[htpb]
  \centering
  \includegraphics[width=0.99\linewidth]{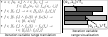}
  \caption{Iteration variable ranges during vloop fusion.}
  \label{fig:fuse_range}
\end{figure}

\noindent\textbf{Variable Loop Fusion:} During compilation, a tensor
compiler infers loop bounds for all operators. In order to do so, the
compiler usually proceeds from the outputs of the operator graph
towards the inputs, inferring the region of a tensor $t$ that needs to
be computed and then using this information to infer the loop bounds
for the operator that computes $t$. As we saw in~\S\ref{sec:fusion},
the application of scheduling transformations such as fusion can lead
to a situation where the variables used in the tensor accesses in an
operator's body are not the same as the loop iteration variables
present after the transformations have been applied. This means that
during bounds inference, one has to repeatedly translate iteration
variable ranges between the transformed and the original
variables. This is straightforward in the case of cloops, but gets
slightly harder in the case of vloop fusion. For the loop nest in
Fig.~\ref{fig:fusion}, Fig~\ref{fig:fuse_range} provides the rules to
translate between the ranges of iteration variables \verb+o+, \verb+i+
and \verb+f+ as well as a visualization of the ranges. Here, $s()$
represents the variable loop bound of the inner loop, while $f_{oif}$,
$f_{fo}$ and $f_{fi}$ represent the relationships between the
variables \verb+o+, \verb+i+ and \verb+f+ such that $f_{fo}(f_0)$ and
$f_{fi}(f_0)$ evaluate to values of \verb+o+ and \verb+i+,
respectively, corresponding to \verb+f+$=f_0$. Similarly,
$f_{oif}(o_0, i_0)$ evaluates to $f_0$\footnote{In the generated code,
as seen in the right pane of Fig.~\ref{fig:fusion}, these functions
take the form of arrays initialized by the prelude. Further, the
computation of the \texttt{f\textsubscript{oif}} array can, in most
cases, be optimized away to only compute the loop bound \texttt{F} of
the fused loop.}.

\noindent\textbf{Named Dimensions:} In~\S\ref{sec:api}, we described
how the user uses named dimensions to specify relationships between
loops as well as tensor dimensions. These dimensions play an important
part in bounds inference as well. Along with the translation between
fused and unfused loop iteration variables described above, one also
needs to translate ranges of variables across producers and consumers
during bounds inference. %% starting from the outputs of the operator
%% graph and moves towards the inputs.
In \Sys, we use named dimensions to easily identify corresponding
iteration variables across such producers and consumers to allow this
translation.
%% This explicit naming of loops and tensor dimensions makes it easy for
%% the user to specify loop bounds as functions of outer loop iteration
%% variables and correspondingly the sizes of slices in tensor dimensions
%% as functions of indices corresponding to outer tensor begin.

\subsection{Storage Access Lowering} \label{sec:lower_storage}
\begin{figure}
  \centering
  \includegraphics[width=0.9\columnwidth]{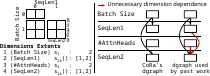}
  \caption{\Sys~precisely models dimension dependences as compared to
    past schemes for sparse tensors.}
  \label{fig:access_lowering}
\end{figure}

In this section, we briefly describe how \Sys~lowers accesses to
ragged tensors. Consider the 4-dimensional attention matrix $X$
involved in a batched implementation of MHA shown in the left pane of
Fig.~\ref{fig:access_lowering}. Here, the first and the third
dimensions are cdims and correspond to the batch size and the number
of attention heads, respectively. The other two dimensions,
corresponding to sequence lengths, are vdims.\footnote{We use the same
layout in \Sys's implementation in~\S\ref{sec:layer_eval}.} For $X$,
the size of a slice for both these vdims is the same function
($s_{24}())$) of the outermost batch dimension. %% In other words,
%% each sequence in the batch has a different length, which is shared
%% across all the attention heads. %

Due to the irregular nature of ragged tensor storage, we need some
auxiliary data structures to be able to lower memory accesses to
$X$. The lowering scheme used by past work on sparse
tensors~\cite{csf, taco_format} assumes that the number of non-zeros
in a slice of a sparse dimension is, in general, a function of all
outer dimensions. However, recall that for our example tensor $X$, the
size of a slice of either vdim depends only on the outermost batch
dimension. Being agnostic to such precise dependencies between tensor
dimensions (as illustrated via the \emph{dimension graphs}, or
\emph{dgraphs} in Fig.~\ref{fig:access_lowering}), past work would
compute and store more auxiliary data as compared to~\Sys.

\Sys's lowering scheme allows for cheap $O(1)$ accesses to ragged
tensors. To enable this, we need to compute a memory offset within a
constant number of operations. The reason sparse tensor formats such
as the CSR format do not allow constant time tensor accesses is
because they explicitly store indices of one or more dimensions along
with every non-zero value. Thus, given a tensor index, one needs to
perform a search over these stored indices to obtain the correct
non-zero element. In the case of ragged tensors, however, we note that
within a vdim slice, the data in densely packed with no intervening
zero elements. Therefore, we can get away without storing explicit
indices for any dimension. Accessing the precomputed memory offsets is
also a constant time operation as \Sys's auxiliary data structures
store these offsets using simple arrays.

We describe these lowering schemes further in the appendix
in~\S\ref{sec:ap_access_lowering}. In short, however, our storage
access lowering scheme reduces the amount of auxiliary data that needs
to be computed thus reducing
%% the memory and computation
overheads of the prelude code (\S\ref{sec:cora_oh_eval}), while
allowing cheap tensor accesses.

%% Due to the irregular nature of ragged tensor storage, we need some
%% auxiliary data structures to be able to lower memory accesses to
%% $X$. The scheme used by past work on sparse tensors~\cite{csf,
%%   taco_format} for this purpose assumes that the number of non-zeros
%% in a slice of a sparse dimension is, in general, a function of all of
%% the outer dimensions. However, recall that for our example tensor $X$,
%% the size of a slice of either vdim depends only on the outermost batch
%% dimension. Being agnostic to such precise dependencies between tensor
%% dimensions (as illustrated via the \emph{dimension graphs}, or
%% \emph{dgraphs} in Fig.~\ref{fig:access_lowering}), past work would
%% compute and store more auxiliary data as compared to~\Sys. \Sys's
%% lowering scheme also exploits the fact the data in a vdim slice is
%% densely packed, which allows $O(1)$ tensor accesses by cleverly
%% precomputing certain memory offsets (the auxiliary data
%% structures). We describe these lowering schemes further in the
%% appendix in~\S\ref{sec:ap_access_lowering}. In short, however, our
%% storage access lowering scheme reduces the amount of auxiliary data
%% that needs to be computed thus reducing the memory and computation
%% overheads of the prelude code (\S\ref{sec:cora_oh_eval}), while
%% allowing cheap tensor accesses.

\section{Implementation}\label{sec:impl}
We prototype \Sys~by extending TVM~\cite{tvm} v0.6, a DL framework and
a tensor compiler. Some details regarding this implementation are
discussed below.

\noindent\textbf{Ragged API: } Our prototype allows vdims to depend on
at most one outer tensor dimension. This is not a fundamental
limitation and can easily be overcome, though we have not needed to
for our evaluation. We implement the operator splitting and hfusion
transforms for non-reduction loops.

\noindent\textbf{Lowering: } Our current prototype does not
auto-schedule the expressed computation. The evaluation therefore uses
implementations optimized using a combination of manual scheduling and
grid search. For some operators, we auto-scheduled the corresponding
dense tensor operator using past work~\cite{ansor} and manually
applied the schedule to the ragged case. We find that this works well
in most cases and therefore believe that the prototype could readily
be extended with prior work on auto-scheduling. Our implementation
currently expects users to correctly allocate memory (taking into
account padding requirements as specified in the schedule) for
tensors. Checks to report these problems can also, however, be easily
implemented.

\section{Evaluation} \label{sec:eval}
We evaluate~\Sys~against state-of-the-art baselines, first, on two
ragged variants of the gemm (general matrix multiplication) operation
in~\S\ref{sec:vtrmm_eval} and then on an encoder layer of the
transformer model (Fig.~\ref{fig:fusion_graph})
in~\S\ref{sec:layer_eval}. Our experimental environment is described
in Table~\ref{table:exp_env}. Below, we refer to the four platforms
listed in the table as Nvidia GPU, Intel CPU, 8-core ARM CPU and
64-core ARM CPU. Our evaluation is performed with single-precision
floating point numbers.

\begin{table}
  \centering \scriptsize
  \vspace{-3mm}
  \caption{Our experimental environment.}
  \vspace{-1mm}
  \begin{threeparttable}[b]
  \addtolength{\tabcolsep}{-3pt}
  %% \begin{tabular}{L{3.7cm}L{3.9cm}}
  \begin{tabular}{L{3.2cm}L{4.5cm}}
    \toprule
    Hardware & Software (All instances ran Ubuntu 20.04.) \\ \midrule
    Nvidia Tesla V100 GPU (Google cloud n1-standard-8 instance) & CUDA 11.1, cuDNN 8.2.1, PyTorch 1.9.0, FasterTransformer v4.0 (commit dd4c071)\\ \midrule
    8 core, 16 thread Intel CascadeLake CPU (Google cloud n2-standard-16 instance) & Intel MKL (v2021.3) \\ \midrule
    8 core ARM Graviton2 CPU (AWS c6g.2xlarge instance) & \multirow{2}{4.5cm}{PyTorch 1.10.0a0+git36449ea (with oneDNN 2.4 and Arm compute library 21.11), TensorFlow 2.6.0 (with oneDNN 2.3 and Arm compute library 21.05), OpenBLAS 0.3.10} \\ \cmidrule(lr){1-1}
    64 core ARM Graviton2 CPU (AWS c6g.16xlarge instance) & \\
    \bottomrule
  \end{tabular}
  \addtolength{\tabcolsep}{3pt}
  \end{threeparttable}
  \label{table:exp_env}
\end{table}

%% \begin{table}
%%   \centering \scriptsize
%%   \vspace{-3mm}
%%   \caption{Experimental environment (GCP: Google Cloud Platform)}
%%   \begin{threeparttable}[b]
%%   \addtolength{\tabcolsep}{-3pt}
%%   %% \begin{tabular}{L{3.7cm}L{3.9cm}}
%%   \begin{tabular}{L{3.2cm}L{4.5cm}}
%%     \toprule
%%     Hardware & Software (All instances ran Ubuntu 20.04.) \\ \midrule
%%     Nvidia Tesla V100 GPU (GCP n1-standard-8 instance) & CUDA 11.1, cuDNN 8.2.1, PyTorch 1.9.0, FasterTransformer v4.0 (commit dd4c071)\\ \midrule
%%     8 core, 16 thread Intel CascadeLake CPU (GCP n2-standard-16 instance) & Intel MKL (v2021.3) \\ \midrule
%%     8 core ARM Graviton2 CPU (AWS c6g.2xlarge instance) & \multirow{2}{4.5cm}{PyTorch 1.10.0a0+git36449ea (with oneDNN 2.4 and Arm compute library 21.11), TensorFlow 2.6.0 (with oneDNN 2.3 and Arm compute library 21.05), OpenBLAS 0.3.10} \\ \cmidrule(lr){1-1}
%%     64 core ARM Graviton2 CPU (AWS c6g.16xlarge instance) & \\
%%     \bottomrule
%%   \end{tabular}
%%   \addtolength{\tabcolsep}{3pt}
%%   \end{threeparttable}
%%   \label{table:exp_env}
%% \end{table}

\subsection{Matrix Multiplication} \label{sec:vtrmm_eval}
%% We start by evaluating~\Sys's performance on variable-sized batched
%% gemm (or vgemm, for short) which involves computing a batch of matrix
%% multiplication operations where the operations all have different
%% dimensions and on triangular matrix multiplication (or trmm, for
%% short), which involves multiplying an upper or lower triangular matrix
%% (we evaluate the lower triangular case, though the upper triangular
%% case is equivalent).

We start by evaluating~\Sys's performance on the variable-sized
batched gemm (or vgemm) and the triangular matrix multiplication (or
trmm) operators. As with all the implementations we compare against,
the \Sys~implementations of these operators use fully padded storage
for all tensors.

\noindent \textbf{Variable-Sized Batched Gemm:} The vgemm operator
consists of a batch of gemm operations, each with different
dimensions. For this operator, we evaluate~\Sys~on the Nvidia GPU and
Intel CPU backends and compare against hand-optimized implementations
of vgemm and fully padded batched gemm in both cases.  On the CPU, we
compare against Intel MKL's implementations while on the GPU, we
compare against past work~\cite{cbt} on vgemm and cuBLAS's
implementation of fully padded batched gemm. We use synthetically
generated data where matrix dimensions are uniformly randomly chosen
multiples of 128 in $[512, 1408]$. \Sys's CPU implementation offloads
the computation of inner gemm tiles to MKL, allowing us to obtain
computational savings due to raggedness while also exploiting MKL's
highly optimized microkernels.

\begin{figure}
  \centering
  \includegraphics[width=0.90\columnwidth]{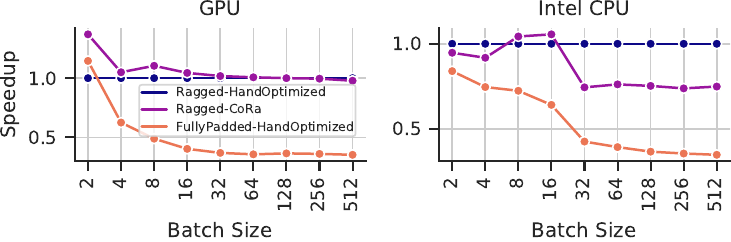}
  %% \caption{\Sys's vgemm performance compared against hand-optimized
  %%   implementations of vgemm and fully padded gemm%% , normalized to the
  %%   %% Ragged-HandOptimized baseline
  %%   .}
  %% \caption{\Sys's vgemm performance compared against hand-optimized
    %% implementations of vgemm and fully padded gemm.}
  \caption{Performance comparison of \Sys's vgemm and hand-optimized
    implementations of vgemm and fully padded gemm.}
  \label{fig:vgemm_eval}
\end{figure}

As Fig.~\ref{fig:vgemm_eval} shows, \Sys~is effectively able to
exploit raggedness on both CPUs and GPUs, performing as well as or
better than the hand-optimized implementation on the GPU and obtaining
better than 73\% of the performance of MKL's vgemm for all batch sizes
and performing better on a couple on the CPU. In all cases, \Sys~is
significantly better than the fully padded gemm operations, which
perform worse at higher batch sizes as there is more wasted computation
as batch size goes up for the batch sizes evaluated.

\noindent \textbf{Triangular Matrix Multiplication:} A triangular
matrix, i.e.~a matrix where all the elements above (or below) the
diagonal are zero, can be thought of as a ragged tensor because all
non-zero elements in a row are densely packed and their number per row
is a function of the row index. Operations on triangular matrices can,
thus, be effectively expressed and optimized using \Sys. In this
section, we evaluate~\Sys~on the trmm operator wherein we multiply a
square lower triangular matrix with a square dense matrix, on the
Nvidia GPU. We compare against cuBLAS's trmm and fully padded gemm
implementations. In trmm, the reduction loop is a vloop. In order to
efficiently handle the last few iterations of this loop after tiling,
we use operation splitting\footnote{HFusion is not applicable here as
the split loop is a reduction loop and executing the split operators
concurrently would require atomic instructions, which our prototype
does not yet support.}  (\S\ref{sec:api}). Further, the raggedness in
this loop leads to imbalanced load across the GPU thread blocks. We
use thread remapping (\S\ref{sec:scheduling}) to schedule thread
blocks with the most amount of work first, leading to more balanced
load.

\begin{figure}
  \centering
  \includegraphics[width=0.90\columnwidth]{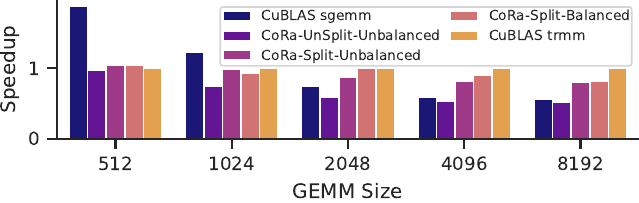}
  \caption{\Sys's trmm performance compared against cuBLAS's
    hand-optimized trmm and fully-padded gemm implementations.}
  \label{fig:trmm_eval}
\end{figure}

Fig.~\ref{fig:trmm_eval} shows the performance of the aforementioned
cuBLAS implementations and three implementations in
\Sys---\Sys-unsplit-unbalanced, \Sys-split-unbalanced and
\Sys-split-balanced---which progressively employ operation splitting
and thread remapping, starting with neither. We see the trmm
implementations---both cuBLAS's and \Sys's---are beneficial as
compared to cuBLAS's dense sgemm operator only for larger matrices. In
all cases, however, the \Sys-split-balanced implementation performs
within 81.3\% of cuBLAS's hand-optimized trmm
implementation. Operation splitting leads to a significant increase in
performance by allowing \Sys~to elide conditional checks in the main
body of the computation. Further, a better load distribution with
thread remapping also helps \Sys~achieve performance close to cuBLAS.

\begin{table}
  \centering
  \scriptsize
  \vspace{-2mm}
  \caption{Datasets used in our evaluation}
  \vspace{-1mm}
  \addtolength{\tabcolsep}{-3pt}
  \begin{tabular}{L{4.7cm}cx{3.3cm}}
    \toprule
    Dataset (Short name, if any)               & Min. / Mean / Max. SeqLength \\ \midrule
    RACE~\cite{race}                           & 80 {} / {} 364 {} / {} 512             \\
    English Wikipedia with SeqLen 512 (Wiki512)& 12 {} / {} 371 {} / {} 512             \\
    SQuAD v2.0~\cite{squadv2} (SQuAD)          & 39 {} / {} 192 {} / {} 384             \\
    English Wikipedia with SeqLen 128 (Wiki128)& 14 {} / {} 117 {} / {} 128             \\
    MNLI~\cite{mnli}                           & 9 {} / {} 43 {} / {} 128               \\
    XNLI~\cite{xnli}                           & 9 {} / {} 70 {} / {} 128               \\
    MRPC~\cite{mrpc}                           & 21 {} / {} 59 {} / {} 102              \\
    CoLA~\cite{cola}                           & 6 {} / {} 13 {} / {} 37                \\
    \bottomrule
  \end{tabular}
  \addtolength{\tabcolsep}{3pt}
  \label{table:datasets}
\end{table}

\subsection{The Transformer Model} \label{sec:layer_eval}
We now move on to look at how \Sys~performs on various modules of the
transformer model. We mainly focus on the GPU backend as it is more
commonly used for these models. %
%% We do, however, evaluate \Sys's performance on a part of the encoder
%% layer of a transformer on the ARM CPU and provide the numbers at the
%% end of this section.
We use a 6 layer model with a hidden dimension of 512 and 8 attention
heads each of size 64. The encoder layer contains two feed-forward
layers, the inner one of which has a dimension of 2048. These are the
same hyperparameters used in the base model evaluated
in~\cite{transformer}. We use sequence lengths from some commonly used
NLP datasets listed in Table~\ref{table:datasets}.\footnote{%% The
%% sequence lengths used correspond to the text obtained after
%% preprocessing as implemented in work on transformer
%% models~\cite{transformer, bert, xlnet}. More details can be found
%% in~\S\ref{sec:ap_datasets}.
More details can be found in \S\ref{sec:ap_datasets}.} We focus on
larger batch sizes (32, 64 and 128) because, as we saw in
Fig.~\ref{fig:flop_ratios}, there is lesser opportunity to exploit
raggedness for smaller batch sizes and hence other factors such as the
quality of the schedules used in \Sys's implementations play a big
role. %% Below, we refer to the
%% multi-head attention and scaled dot-product attention sub-modules of
%% the transformer model as MHA and SDPA for short.
In this section, \Sys's~implementations use ragged tensor storage.

\begin{table}
  \centering
  \scriptsize
  \vspace{-2mm}
  \caption{Transformer encoder layer execution latencies (in ms) for
    \Sys, PyTorch and the two manually-optimized variants of
    FasterTransformer on the Nvidia GPU. \Sys's execution latencies
    include prelude overheads assuming a 6 layer transformer encoder.}
  \addtolength{\tabcolsep}{-2pt}
  \begin{tabular}{cx{1.2cm} cx{1cm} cx{1.5cm} cx{1.5cm} cx{1.5cm} cx{1.5cm}}
    \toprule
    Dataset &  Batch Size & PyTorch & FT   & \Sys      & FT-Eff         \\ \midrule
    \multirow{3}{*}{RACE} & 32 & 12.22 & 11.0 & 8.22 & 8.61 \\
 & 64 & 24.46 & 21.88 & 15.91 & 16.75 \\
 & 128 & 48.73 & 42.26 & 31.45 & 33.61 \\ \hline
\multirow{3}{*}{Wiki512} & 32 & 12.26 & 11.0 & 9.1 & 9.32 \\
 & 64 & 24.52 & 22.12 & 17.4 & 17.85 \\
 & 128 & 48.72 & 42.43 & 32.17 & 33.66 \\ \hline
\multirow{3}{*}{SQuAD} & 32 & 8.17 & 7.56 & 4.15 & 4.69 \\
 & 64 & 16.9 & 15.63 & 7.78 & 9.2 \\
 & 128 & 34.18 & 30.62 & 15.36 & 17.91 \\ \hline
\multirow{3}{*}{Wiki128} & 32 & 2.79 & 2.45 & 2.59 & 2.28 \\
 & 64 & 5.12 & 4.61 & 4.72 & 4.35 \\
 & 128 & 10.1 & 9.29 & 8.86 & 8.54 \\ \hline
\multirow{3}{*}{MNLI} & 32 & 2.22 & 2.04 & 1.11 & 1.03 \\
 & 64 & 4.44 & 4.06 & 1.89 & 1.93 \\
 & 128 & 9.53 & 8.86 & 3.53 & 3.78 \\ \hline
\multirow{3}{*}{XNLI} & 32 & 2.76 & 2.45 & 1.56 & 1.5 \\
 & 64 & 5.13 & 4.62 & 2.94 & 2.86 \\
 & 128 & 10.03 & 9.3 & 5.62 & 5.49 \\ \hline
\multirow{3}{*}{MRPC} & 32 & 1.85 & 1.73 & 1.32 & 1.27 \\
 & 64 & 3.76 & 3.48 & 2.6 & 2.36 \\
 & 128 & 7.42 & 6.89 & 4.55 & 4.55 \\ \hline
\multirow{3}{*}{CoLA} & 32 & 0.67 & 0.57 & 0.59 & 0.44 \\
 & 64 & 1.02 & 0.93 & 0.77 & 0.63 \\
 & 128 & 2.37 & 2.18 & 1.26 & 1.17 \\ \bottomrule

  \end{tabular}
  \addtolength{\tabcolsep}{2pt}
  \label{table:layer_eval}
  \vspace{-1mm}
\end{table}

\noindent\textbf{Transformer Encoder Layer:} We first evaluate the
forward pass latency of an encoder layer of the transformer model
(Fig.~\ref{fig:fusion_graph}). We compare \Sys's performance with that
of FasterTransformer and an implementation in PyTorch, a popular DL
framework, with TorchScript~\cite{TorchScript} enabled. %
%% The transformer model contains a number of position-wise feed-forward
%% layers, both as part of the MHA module (Proj1 and Proj2 in
%% Fig.~\ref{fig:fusion_graph}) as well as the fully connected
%% layers. Because these operations project the hidden vectors associated
%% with each word independently, with manual effort, they can be
%% implemented as gemm operators without any padding.
All the operators in the encoder layer except the ones in the SDPA
sub-module process the hidden vectors associated with each word
independently. Therefore, with manual effort, they can be implemented
without any padding. The linear transformation operators Proj1, Proj2,
FF1 and FF2 reduce to gemm operations in this case. FasterTransformer
provides an option to perform this optimization, first introduced in
EffectiveTransformers~\cite{EffectiveTrans}. We compare against
FasterTransformer both with and without this optimization. We refer to
these two implementations as FT-Eff and FT, respectively. In the
\Sys~implementation, this optimization is applied simply by loop
fusion, analogous to the illustration in Fig.~\ref{fig:fusion}. In
\Sys's implementation however, we pad this fused loop so that its
bound is a multiple of 64. In other words, we add a padding sequence
to the batch to ensure that the sum of the sequence lengths is a
multiple of 64. We refer to this kind of padding as \emph{bulk
padding} (Fig.~\ref{fig:fusion_graph}). The relative amount of bulk
padding added is usually quite low as the sum of sequence lengths in a
batch is much higher.

\begin{figure}
\centering
\begin{minipage}{.48\columnwidth}
  \centering
  \includegraphics[width=1.0\columnwidth]{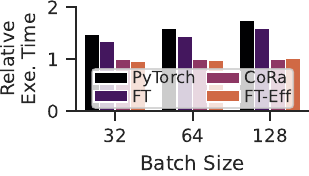}
  \captionof{figure}{Relative execution times of the transformer
    encoder layer on the GPU.}
  %% \captionof{figure}{Relative GPU execution times for PyTorch,
    %% FasterTransformer and \Sys~for the transformer encoder layer.}
  \label{fig:layer_summary}
  \vspace{-1mm}
\end{minipage}\hfill%
\begin{minipage}{.48\columnwidth}
  \centering
  \includegraphics[width=1.0\columnwidth]{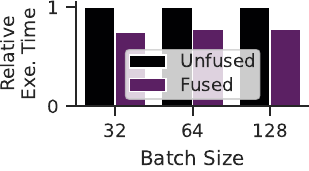}
  %% \captionof{figure}{Relative MHA execution times with and without
    %% layout change operator fusion.}
  \captionof{figure}{Benefits of padding change operator fusion for
    the MHA module.}
  \label{fig:pad_fusion_eval}
  \vspace{-1mm}
\end{minipage}
\end{figure}

Table~\ref{table:layer_eval} shows the forward execution latencies for
the encoder layer for the aforementioned frameworks and datasets. The
auxiliary data structures computed by \Sys's prelude are shared across
multiple layers of the model as the raggedness pattern stays the same
across layers, depending only on the sequence lengths in the
mini-batch. The execution times shown for \Sys~include per-layer
prelude overheads assuming a 6 layer model. We further look at these
overheads in~\S\ref{sec:cora_oh_eval}. As we can see, the
\Sys~implementation is competitive with the manually-optimized FT-Eff
implementation for all datasets, even performing better in a few
cases, and performs significantly better as compared to the
fully-padded PyTorch and FT
implementations. Fig.~\ref{fig:layer_summary}, which plots the overall
performance of all these implementations for the batch sizes
evaluated, makes this clear.

\begin{figure}
  \centering
  \includegraphics[width=0.95\columnwidth]{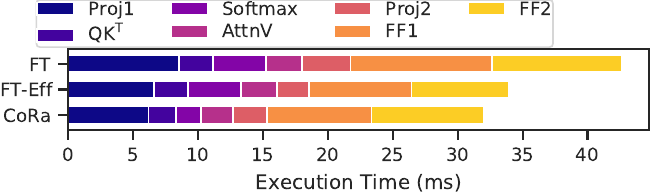}
  \caption{Breakdown of the encoder layer execution times for the RACE
    dataset at batch size 128. This data is obtained with profiling
    turned on and might deviate from Table~\ref{table:layer_eval}.}
  \label{fig:per_ops_plot}
\end{figure}

We now take a closer look at the FasterTransformer and
\Sys~implementations which are sketched in
Fig.~\ref{fig:fusion_graph}.\footnote{FasterTransformer uses
specialized implementations for different GPUs. In this paper, we
limit our discussion to its implementation for the Nvidia V100 GPU we
use for evaluation.} The FT implementation is similar to the FT-Eff
implementation except it uses full padding for all operations. The
\Sys~and FasterTransformer implementations differ in their operator
fusion strategies. Therefore, the figure breaks the implementations
down to the smallest sub-graphs that correspond to each
other. Fig.~\ref{fig:per_ops_plot} shows a breakdown of the execution
times for these implementations for the RACE dataset and batch size
128 at the level of these sub-graphs.\footnote{The raw data for this
plot is listed in Table~\ref{table:ap_op_times} in the appendix.} As
Fig.~\ref{fig:fusion_graph} shows, the FT-Eff and \Sys~implementations
differ significantly with respect to padding only in the SDPA
sub-module where the FT-Eff implementation employs full padding while
the \Sys~employs partial padding. We see, in
Fig~\ref{fig:per_ops_plot}, that the \Sys~implementation performs
better than FasterTransformer for all the SDPA operators
(QK\textsuperscript{T}, Softmax and AttnV) despite the fact that the
latter is heavily hand optimized.\footnote{The execution times of the
three SDPA operators is quadratically proportional to the sequence
length, unlike the remaining operators which are linearly
proportional. We discuss the performance of SDPA further
in~\S\ref{sec:ap_breakdown} of the appendix.} This is because \Sys's
ability to handle raggedness enables it to perform less wasted
computation. For the remaining operators where both the \Sys~and
FT-Eff implementations employ little to no padding, we see that the
\Sys~implementation is usually slower, but often close in performance
to the FT-Eff implementation and significantly faster than the fully
padded FT implementation. This is expected as FT-Eff calls into
cuBLAS's extensively optimized gemm kernels for the linear
transformation operators and into hand-optimized kernels for the
rest. \Sys's performance drops slightly for datasets with smaller
sequence lengths as well as for smaller batch sizes. As we discuss
in~\S\ref{sec:ap_breakdown}, this performance difference can be
reduced by further optimizing the schedules used for the projection
and feed forward operators in \Sys's implementation for smaller batch
sizes and sequence lengths. Further, we also note that the overheads
associated with the prelude code and partial padding
(\S\ref{sec:cora_oh_eval}) play a larger role in these cases, further
contributing to increased execution latencies.

FasterTransformer's reliance on vendor libraries prevents it from
fusing any of the gemm operations with surrounding elementwise
operators, which \Sys~can due to its compiler-based
approach. Specifically \Sys~can completely fuse all operators which
add or remove padding in its implementation (as shown in
Fig.~\ref{fig:fusion_graph}). This is as opposed to the FT-Eff
implementation, which cannot. Fusing these padding change operators
leads to a significant drop in \Sys's execution latency as seen in
Fig.~\ref{fig:pad_fusion_eval}, which shows the execution latencies of
the MHA module for the RACE dataset in \Sys~with and without this
fusion enabled.

%% \begin{figure}
  %% \centering
  %% \includegraphics[width=0.45\columnwidth]{fig/pad_fusion_eval}
  %% \caption{Execution latencies of the \Sys~implemention of the MHA
    %% module with and without the layout change operators fused.}
  %% \label{fig:pad_fusion_eval}
%% \end{figure}

%% \begin{wrapfigure}{r}{0.45\columnwidth}
%%   \begin{center}
%%     \includegraphics[width=0.45\columnwidth]{fig/layer_summary_eval}
%%   \end{center}
%%   %% \caption{Relative execution times for PyTorch, FasterTransformer and
%%     %% \Sys~performance on the transformer encoder layer on the GPU
%%     %% backend.}
%%   \label{fig:summary_eval}
%% \end{wrapfigure}

\noindent\textbf{Masked Scaled Dot-Product Attention:} The decoder
layer of a transformer uses a variant of MHA called \textit{masked MHA} wherein
the upper half of the attention matrix is masked for all attention
heads during training. This masking only affects the SDPA module, the
operators in which can now be seen as computing on a batch of lower
triangular matrices. We saw in~\S\ref{sec:vtrmm_eval} that \Sys~can
effectively generate code for operations on triangular matrices. For
batch size 128, an implementation of masked SDPA in \Sys~which
exploits this masking performs $1.56\times$ faster than an
implementation which does not for the RACE dataset and $1.29\times$
for the MNLI dataset. The benefits are less pronounced for the MNLI
dataset, which has smaller sequence lengths, as we pad vloops to be
multiples of a constant regardless of the dataset. We provide more
data and discussion on the implementation of masked SDPA
in~\S\ref{sec:ap_mmha_eval} in the appendix.

%% \begin{figure}
  %% \centering
  %% \includegraphics[width=0.90\columnwidth]{fig/mmha_eval}
  %% \caption{Execution time of the masked scaled dot-product attention
    %% in PyTorch and \Sys, with and without padding the triangular
    %% attention matrix.}
  %% \label{fig:mmha_eval}
%% \end{figure}

%% \noindent\textbf{Memory Consumption:} Along with the computational
%% benefits of ragged tensors as demonstrated above, their use in
%% transformer models can also lead to a non-trivial drop in the
%% memory usage as is seen in Fig.~\ref{fig:mem_eval}. In the figure,
%% we compare the peak memory usage (in MBs) for the forward pass of
%% one transformer encoder layer with the same hyperparameters as
%% above for all the datasets in Table~\ref{table:datasets} for a
%% batch size of 32. While in the evaluation above, both the \Sys~and
%% FasterTransformer implementations preallocate memory,
%% Fig.~\ref{fig:mem_eval} shows the peak memory usage obtained from
%% simulations of the case where memory is allocated and freed
%% aggressively. Both FT and FT-Eff allocate memory for the fully
%% padded case and hence we only show data for that case. As can be
%% seen, \Sys's use of ragged tensor storage leads to a 17\% drop in
%% the memory usage as compared to the fully padded FasterTransformer
%% implementation. When training, this drop in memory usage can
%% translate to a corresponding beneficial rise in the mini batch
%% size.

\noindent\textbf{Memory Consumption:} We find that the use of ragged
tensors leads to an overall $1.78\times$ drop in the size of the
forward activations (computed analytically) of the encoder layer
across all datasets at batch size 64 (more details
in~\S\ref{sec:ap_mem}). The reduction, however, is not uniform across
the datasets and those with higher mean sequence lengths, such as
Wiki512 and Wiki128, see only small benefits. Forward activations
often consume significant memory during training and ragged tensors
can help alleviate memory bottlenecks along with other memory
management techniques for training~\cite{dtr, checkmate}.

%% \noindent\textbf{Memory Consumption:} Along with the computational
%% benefits of ragged tensors, their use can also lower memory
%% usage. Fig.~\ref{fig:mem_eval} shows, for datasets in
%% Table~\ref{table:datasets} and a batch size of 64, the relative total
%% memory consumption (computed analytically) of the forward activations
%% of a transformer encoder layer for \Sys's implementation with and
%% without the use of ragged tensors. %
%% %% We take into account any padding that the ragged implementation
%% %% needs. The relative memory consumption for the other batch sizes is
%% %% also similar.
%% %% The memory consumption in both cases is computed analytically and
%% %% excludes model weights.
%% Overall, we see a $1.78\times$ drop in memory usage. The reduction,
%% however, is not uniform across the datasets. Only small improvements
%% are observed for the Wiki512 and Wiki128 datasets, which have higher
%% sequence lengths on average (seen in
%% Table~\ref{table:datasets}). Forward activations often consume
%% significant memory during training and ragged tensors can help
%% alleviate memory bottlenecks along with with other memory management
%% techniques for training~\cite{dtr, checkmate}.

\begin{table}
  \centering \scriptsize
  \vspace{-2mm}
  \caption{MHA execution latencies (in ms) on the 64-core ARM CPU for
    TensorFlow and \Sys.}
  \addtolength{\tabcolsep}{-4pt}
  \resizebox{0.98\columnwidth}{!}{%
    \begin{tabular}{cx{0.8cm} %
        cx{0.8cm} cx{0.8cm} cx{0.8cm} %@{\hspace{0.1\tabcolsep}} %
        cx{0.8cm} cx{0.8cm} cx{0.8cm} %@{\hspace{0.1\tabcolsep}} %
        cx{0.8cm} cx{0.8cm} cx{0.8cm} %@{\hspace{0.1\tabcolsep}} %
      }
      \toprule
      \multirow{2}{0.8cm}{\centering Dataset} & \multicolumn{3}{c}{Batch Size 32} & \multicolumn{3}{c}{Batch Size 64} & \multicolumn{3}{c}{Batch Size 128} \\ \cmidrule(lr){2-10}
      & TF & TF-UB & \Sys                    & TF & TF-UB & \Sys                    & TF & TF-UB & \Sys                     \\ \midrule
      RACE & 55 & 46 & \textbf{44} & 111 & 88 & \textbf{85} & 209 & \textbf{156} & 168 \\
Wiki512 & 53 & 53 & \textbf{47} & 106 & 96 & \textbf{91} & 205 & \textbf{172} & 176 \\
SQuAD & 35 & 27 & \textbf{20} & 68 & 49 & \textbf{39} & 137 & 79 & \textbf{76} \\
Wiki128 & 11 & 11 & \textbf{9} & 19 & 18 & \textbf{17} & 34 & 33 & \textbf{33} \\
MNLI & 9 & 9 & \textbf{4} & 16 & 14 & \textbf{7} & 30 & 23 & \textbf{14} \\
XNLI & 11 & 11 & \textbf{6} & 18 & 18 & \textbf{11} & 34 & 28 & \textbf{22} \\
MRPC & 9 & 8 & \textbf{5} & 14 & 14 & \textbf{10} & 26 & 23 & \textbf{18} \\
CoLA & 5 & 4 & \textbf{2} & 6 & 6 & \textbf{3} & 9 & 8 & \textbf{5} \\ \bottomrule

    \end{tabular}
  }
  \addtolength{\tabcolsep}{4pt}
  \label{table:arm_layer_eval}
  \vspace{-1mm}
\end{table}

\noindent\textbf{MHA Evaluation on ARM CPU:}
Table~\ref{table:arm_layer_eval} shows the execution latencies of MHA
implementations in TensorFlow and \Sys~on the 64-core ARM CPU. We
evaluate against two execution configurations of TensorFlow---TF,
where the entire mini-batch is executed at once and TF-UB, where the
mini-batch is executed as a series of smaller \emph{micro-batches},
which enables execution with lower padding. Across the datasets and
batch sizes evaluated, we see that \Sys's implementation is overall
$1.57\times$ faster than TF and $1.37\times$ faster than TF-UB. In
this case, too, \Sys's ability to save on wasted computation due to
padding leads to significant performance gains over a popular DL
framework. \S\ref{sec:ap_breakdown} of the appendix more extensively
compares the performance of TensorFlow and PyTorch against \Sys~on
both the 8- and 64-core ARM CPUs.

%% Table~\ref{table:arm_layer_eval} shows the execution latencies for the
%% MHA module in PyTorch, TensorFlow and \Sys~on the ARM CPU
%% backend. Across the datasets and batch sizes evaluated, we see that
%% \Sys's implementation is overall $1.86\times$ faster than PyTorch and
%% $1.89\times$ faster than TensorFlow. In this case, too, \Sys's ability
%% to save on wasted computation due to padding leads to significant
%% performance gains over popular DL frameworks. We provide more details
%% and discussion in~\S\ref{sec:ap_breakdown} of the appendix.

%% \begin{table}
%%   \centering \scriptsize
%%   \vspace{-3.5mm}
%%   \caption{MHA execution latencies (in ms) on the ARM CPU. TF and PT
%%     refer to TensorFlow and PyTorch respectively.}
%%   \addtolength{\tabcolsep}{-4pt} \resizebox{0.98\columnwidth}{!}{%
%%     \begin{tabular}{cx{0.8cm} %
%%         cx{0.8cm} cx{0.8cm} cx{0.8cm} %@{\hspace{0.1\tabcolsep}} %
%%         cx{0.8cm} cx{0.8cm} cx{0.8cm} %@{\hspace{0.1\tabcolsep}} %
%%         cx{0.8cm} cx{0.8cm} cx{0.8cm} %@{\hspace{0.1\tabcolsep}} %
%%       }
%%       \toprule
%%       \multirow{2}{0.8cm}{\centering Dataset} & \multicolumn{3}{c}{Batch Size 32} & \multicolumn{3}{c}{Batch Size 64} & \multicolumn{3}{c}{Batch Size 128} \\ \cmidrule(lr){2-10}
%%       & PT & TF & \Sys                    & PT & TF & \Sys                    & PT & TF & \Sys                     \\ \midrule
%%       \input{src/bert_layer_table_cpu.tex}
%%     \end{tabular}
%%   }
%%   \addtolength{\tabcolsep}{4pt}
%%   \label{table:arm_layer_eval}
%%   \vspace{-2.5mm}
%% \end{table}

\begin{figure}
  \centering
  \includegraphics[width=0.92\columnwidth]{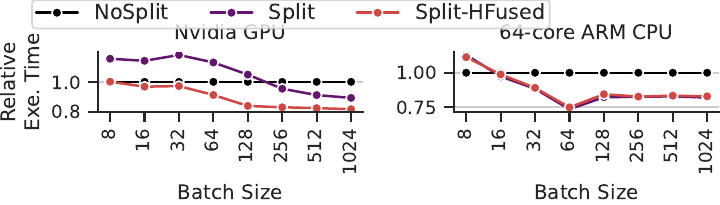}
  \caption{Benefits of operator splitting and hfusion. Note that the
    y-axis does not start at 0.}
  \label{fig:bin_packed_eval}
  \vspace{-1mm}
\end{figure}

\subsection{Operation Splitting and Horizontal Fusion}\label{sec:bp_eval}
We now evaluate operator splitting and hfusion on the AttnV operator,
which is an instance of the vgemm problem. AttnV has two vloops, one
of which is a reduction loop. We apply the optimizations to the
non-reduction vloop allowing us to use a larger tile size (we use 64)
without padding the vloop bound to be a multiple of this tile
size. This especially benefits datasets with sequence lengths
comparable to the tile size, such as MNLI. For this dataset,
Fig.~\ref{fig:bin_packed_eval} shows the relative execution times of
three \Sys~implementations of AttnV---NoSplit, Split and
Split-HFused---in which we progressively perform the two
optimizations, on the Nvidia GPU and 64-core ARM CPU backends. On the
GPU, operation splitting causes a slowdown despite lower wasted
computation as it reduces parallelism, which is restored by
hfusion. This is more apparent at lower batch sizes when the amount of
parallelism is lower. The effects of reduced parallelism due to
operation splitting are less apparent on the CPU as it exposes lower
levels of hardware parallelism. The lower levels of CPU parallelism
also mean that hfusion has no benefit in this case. We also evaluate
these optimizations on the QK\textsuperscript{T} operator
in~\S\ref{sec:ap_bp} in the appendix.

\subsection{Overheads in \Sys} \label{sec:cora_oh_eval}
Let us now look at the sources of overheads in \Sys---the prelude
code, the wasted computation due to partial padding and auxiliary data
structure accesses in the generated code.

\noindent\textbf{Prelude Overheads:} The prelude code constructs the
required auxiliary data structures (\S\ref{sec:lowering}) and copies
them to the accelerator's memory if needed. The table below lists the
execution time (in ms) and memory (in kB) required for these tasks for
a 6-layer transformer encoder on the GPU backend. It also shows the
overheads associated with the storage lowering scheme used in past
work we discussed in~\S\ref{sec:lower_storage} (referred to as Sparse
Storage in the table). As compared to this scheme, we see that \Sys's
specialized lowering scheme significantly reduces the resources
required to compute the data structures associated with tensor
storage. The overheads associated with loop fusion are higher than
those associated with storage as we need to compute and store the
relationship between all values of the fused and unfused loop
iteration variables (\S\ref{sec:fusion}). Copying the generated data
structures to the GPU's memory is, however, the major source of the
overhead. Overall, the overheads range from 0.7\% (RACE dataset at
batch size 128) to about 7\% (CoLA dataset at batch size 32) of the
total execution time of the encoder layer on the GPU. On the CPU, the
overheads are a very small fraction of the execution times, because
the execution times are much higher and because the memory copy costs
are absent. We discuss some simple optimizations to reduce prelude
overheads in~\S\ref{sec:ap_cora_overheads} of the appendix.

\begin{table}[H]
  \centering \resizebox{0.99\columnwidth}{!}{%
\begin{tabular}{cccccccc}
  \toprule
  \multirow{2}{1.5cm}{\centering Dataset / Batch Size} & Sparse Storage & \Sys~Storage & \Sys~Loop Fusion & \multirow{2}{1.6cm}{\centering \Sys-Copy Time}\\
                                  &   Time / Mem.           &   Time / Mem.              &     Time / Mem.   &   \\ \midrule
  CoLA / 32 & 0.09 / 267.97 & 3.80e-03 / 2.93 & 5.35e-03 / 32.15 & 0.24 \\
CoLA / 128 & 0.35 / 1047.22 & 5.76e-03 / 11.18 & 0.02 / 104.22 & 0.27 \\
RACE / 32 & 0.52 / 1607.97 & 4.15e-03 / 2.93 & 0.09 / 666.54 & 0.42 \\
RACE / 128 & 2.02 / 6300.02 & 6.30e-03 / 11.18 & 0.34 / 2609.58 & 0.99 \\ \bottomrule

\end{tabular}
}
  \vspace{-4mm}
\end{table}

\noindent\textbf{Partial Padding Overheads:} We saw that in \Sys,
small amounts of padding can be specified for vloops (both unfused
vloops and fused ones with bulk padding) and tensor storage to enable
efficient code generation. While this leads to some wasted
computation, we find that it is generally quite low. For the
transformer encoder layer, we see a 3.5\% increase in the amount of
computation (computed analytically) over the ideal case without
padding for a batch size of 32 and a 2.3\% increase for a batch size
of 128 across all the datasets evaluated. The overheads decrease with
increasing batch size as bulk padding ensures that the sum of the
sequence lengths in a batch is a multiple of a constant (64, in this
case) irrespective of the batch size leading to a higher relative
amount of padding at lower batch sizes. We provide further data and
discussion in~\S\ref{sec:ap_cora_overheads} of the appendix.

%% The overheads decrease with increasing batch size as we employ bulk
%% padding in \Sys's implementation of the encoder layer as discussed
%% above. Because we ensure that the sum of the sequence lengths in a
%% batch is a multiple of a constant (64, in this case) irrespective of
%% the batch size, the relative amount of padding added is higher at
%% lower batch sizes. We provide further data and discussion
%% in~\S\ref{sec:ap_cora_overheads} of the appendix.

\noindent\textbf{Ragged Tensor Overheads and Load Hoisting:} \Sys's
generated code accesses the auxiliary data structures generated by the
prelude leading to frequent indirect memory accesses. We measure the
overheads caused by these accesses for the operators used in
MHA. While the data and more discussion are provided
in~\S\ref{sec:ap_cora_overheads}, we note here that the indirect
memory accesses do not cause any significant slowdown for the Proj1,
Softmax, AttnV and the Proj2 operators. The accesses do lead to a
higher slowdown in the QK\textsuperscript{T} operator, which is the
only operator where we fuse two vloops leading to complex memory
access expressions. For this case, we find that hoisting data
structures accesses outside loops when possible helps recover the lost
performance.

\subsection{Evaluation Against Sparse Tensor Compilers}\label{sec:taco_eval}
We saw that ragged tensors are similar to sparse tensors as both
involve irregular storage. In order to evaluate the suitability of
using sparse tensor compilers for ragged tensors, we compared \Sys's
performance with Taco, a state-of-the-art sparse tensor compiler on a
few triangular matrix operators (implemented in Taco using the CSR and
blocked CSR formats). These implementations perform $1.33\times$ to
$95.37\times$ slower than the corresponding \Sys~implementations.
While \S\ref{sec:ap_taco_eval} of the appendix provides further
details and discussion, we note here that this is essentially due to a
mismatch between the use case of ragged tensors and the general sparse
tensor computations that Taco is designed for. For example, ragged
tensors are usually much denser as compared to traditionally used
sparse tensors and the applications each is used in are quite
different.

%% We saw that ragged tensors are similar to sparse tensors as both
%% involve irregular storage. In order to evaluate the suitability of
%% using sparse tensor compilers for ragged tensors, we compared
%% \Sys's performance with Taco, a state-of-the-art sparse tensor
%% compiler. Specifically, we measured the performance of a few
%% operators on triangular matrices implemented in Taco using the CSR
%% and blocked CSR formats. We provide more detailed discussion
%% in~\S\ref{sec:ap_taco_eval}, but note here that these
%% implementations showed slowdowns ranging from $1.33\times$ to
%% $95.37\times$ compared to the corresponding
%% \Sys~implementations. As we discuss later, this is essentially due
%% to a mismatch between the use case of ragged tensors and the
%% general sparse tensor computations that Taco is designed for. For
%% example, ragged tensors are usually much denser as compared to
%% traditionally used sparse tensors and the applications each is used
%% in are quite different.

\section{Related Work} \label{sec:related}
\noindent\textbf{Tensor Compilers:} There has been extensive work on
tensor compilers such as TVM~\cite{tvm}, Halide~\cite{halide},
Tiramisu~\cite{tiramisu}, Tensor Comprehensions~\cite{tc},
Fireiron~\cite{fireiron}, Stripe~\cite{stripe}, AKG~\cite{akg} as well
as work by~\cite{climate_comp} and~\cite{tensor_core_tc} for dense
tensors and Taco~\cite{taco}, COMET~\cite{comet}, and work by
~\cite{mlir_sparse} and~\cite{sparse_array} for sparse tensors. This
work has informed \Sys's design. We generalize the abstractions
provided by dense tensor compilers to ragged tensors, while enabling
efficient code generation for the latter. We discuss in
\S\ref{sec:taco_eval} and further in \S\ref{sec:ap_taco_eval}, how
despite the similarity between ragged and sparse tensors, sparse
tensor compilers are unable to effectively exploit the properties of
ragged tensors to enable efficient execution.

Past work on DL compilers has also looked at handling
dynamism. Nimble~\cite{nimble} develops dynamism-aware compiler
abstractions from the ground up. Its handling of shape dynamism is
limited to variation across mini-batches. \Sys~is therefore
complementary to Nimble's techniques. Cortex~\cite{cortex} handles
recursive models by essentially lowering the recursive control flow
into sequential control flow on ragged tensors. \Sys~can therefore
potentially be used as part of its pipeline. \Sys's use of
uninterpreted functions and named dimensions has been inspired by
their use in Cortex and past work on the Sparse Polyhedral
Framework~\cite{spf1, spf2, spf3}. Named dimensions are also similar
to the index labels in COMET. \Sys~implements a limited form of the
hfusion optimization, first proposed in~\cite{hfusion}, as part of a
tensor compiler.

\noindent\textbf{DL Frameworks and Graph Optimizations:} DL frameworks
have recently begun adding support for ragged tensors with the
RaggedTensor~\cite{TFRagged} class in TensorFlow and the
NestedTensor~\cite{PTNested} module for PyTorch. Very few operators
are, however, supported for ragged tensors at this
point~\cite{TFIssue, PTIssue}.\footnote{Tensor contraction and similar
operators such as batched gemm and convolution are generally not
supported. PyTorch's NestedTensor further supports only a few
elementwise and reduction operators~\cite{NTL} at this point.}
\Sys~can be used to expand the set of ragged operators supported in
these frameworks. \Sys's techniques are complementary to graph
optimizations for efficient DL execution such as data layout
optimizations~\cite{layout_opt}, kernel fusion~\cite{fusion} and
operator scheduling~\cite{ios} and can be used in conjunction with
them.

\noindent\textbf{Hand Optimized Implementations:} There has been work
on efficient implementations of certain important ragged tensor
operations. This includes the work on variable-sized batched gemm
operations~\cite{cbt, magma} as well as the work on Effective
Transformers and FasterTransformers, which we compared~\Sys's
performance against in~\S\ref{sec:eval}. This past work informs our
work on~\Sys~as we saw with the operator splitting transform
in~\S\ref{sec:scheduling}.

\noindent\textbf{Sparse Tensor Algebra:} There has been decades of
past on work on efficient execution of sparse tensor operators. This
work has been revisited recently in the context of DL by work on
exploiting block sparsity in model weights~\cite{block_sparse} as well
as for tuning sparse kernels for the sparsity patterns and
distributions usually encountered in DL~\cite{dl_sparse}. The thread
remapping strategy discussed in~\S\ref{sec:scheduling} was implemented
first in~\cite{dl_sparse}.

\section{Conclusion} \label{sec:conclusion}
This paper presented \Sys, a tensor compiler for expressing and
optimizing ragged operators to portably target CPUs and GPUs using
simple and familiar abstractions. \Sys's approach, specialized for
ragged tensors, reduces overheads associated with techniques such as
masking and padding. With DL being applied to an ever-increasing set
of fields and the models getting more resource-intensive, we believe
that efficiently handling the shape dynamism that naturally arises in
many settings is important. \Sys{} extends past work on tensor
compilers by supporting efficient operators on ragged tensors. Our
work can also be seen as a step towards unifying past work on sparse
and dense tensor compilation. In the future, we plan to make
\Sys~easier to use, potentially with the help of auto-scheduling
techniques.%, and then to release \Sys~open source.

\section*{Acknowledgments}
This work was supported in part by grants from the National Science
Foundation, Oracle and IBM, and by the Parallel Data Lab (PDL)
Consortium (Amazon, Facebook, Google, Hewlett-Packard Enterprise,
Hitachi, IBM, Intel, Microsoft, NetApp, Oracle, Pure Storage,
Salesforce, Samsung, Seagate, TwoSigma and Western Digital). We would
like to thank Saman Amarasinghe, Dominic Chen, Stephen Chou, Chris
Fallin, Graham Neubig, Olatunji Ruwase, the Catalyst Research Group at
Carnegie Mellon University as well as the anonymous reviewers at MLSys
for their valuable suggestions and feedback on our work.

\bibliography{paper}
\bibliographystyle{mlsys2022}

%%%%%%%%%%%%%%%%%%%%%%%%%%%%%%%%%%%%%%%%%%%%%%%%%%%%%%%%%%%%%%%%%%%%%%%%%%%%%%%
%%%%%%%%%%%%%%%%%%%%%%%%%%%%%%%%%%%%%%%%%%%%%%%%%%%%%%%%%%%%%%%%%%%%%%%%%%%%%%%
% SUPPLEMENTAL CONTENT AS APPENDIX AFTER REFERENCES
%%%%%%%%%%%%%%%%%%%%%%%%%%%%%%%%%%%%%%%%%%%%%%%%%%%%%%%%%%%%%%%%%%%%%%%%%%%%%%%
%%%%%%%%%%%%%%%%%%%%%%%%%%%%%%%%%%%%%%%%%%%%%%%%%%%%%%%%%%%%%%%%%%%%%%%%%%%%%%%
\clearpage
\appendix
We now look at additional details regarding \Sys's~mechanism
in~\S\ref{sec:ap_api},~\S\ref{sec:ap_lowering}
and~\S\ref{sec:ap_impl}, and discuss further aspects of the evaluation
in~\S\ref{sec:ap_eval}. Notably, we look at how \Sys~can exploit
masking in masked MHA to obtain further savings
in~\S\ref{sec:ap_mmha_eval}, discuss how \Sys's overheads are quite
low, allowing it to effectively exploit raggedness
(\S\ref{sec:ap_cora_overheads}) and look more closely at \Sys's
performance on the transformer model and where the benefits come from
in~\S\ref{sec:ap_breakdown}.

\section{Ragged API} \label{sec:ap_api}
\subsection{Thread Remapping Policy} \label{sec:ap_remap}
\begin{figure}[htpb]
  \centering
  \includegraphics[width=0.9\columnwidth]{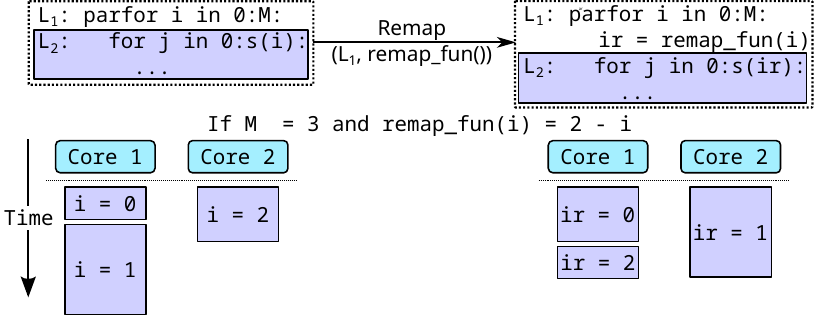}
  \caption{Thread remapping allows users to influence the scheduling
    of iterations to allow for better load balancing.}
  \label{fig:ap_thread_remap}
\end{figure}

We discussed, in~\S\ref{sec:api}, that \Sys~allows users to specify a
thread remapping policy to influence how iterations of a parallel loop
are scheduled on the execution units in the hardware substrate. This
is illustrated in Fig~\ref{fig:ap_thread_remap}.

\section{Ragged API Lowering} \label{sec:ap_lowering}
\subsection{Tensor Storage Lowering} \label{sec:ap_access_lowering}
\begin{figure*}
  \centering
  \includegraphics[width=0.95\linewidth]{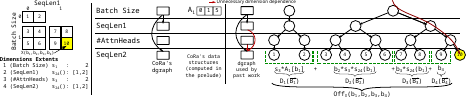}
  \caption{Comparing \Sys's storage lowering with the tree-based
    scheme used by past work on sparse tensors.}
  \label{fig:access_lowering_large}
\end{figure*}

In~\S\ref{sec:lower_storage}, we briefly discussed the storage
lowering schemes used by past work on sparse tensor compilers and by
\Sys. Both are illustrated in Fig.~\ref{fig:access_lowering_large} and
discussed more below.

\noindent\textbf{Sparse Storage Access Lowering Scheme Used in Past
  Work:} Recall the 4-dimensional attention tensor $X$ we discussed
in~\S\ref{sec:lower_storage} and which is illustrated again in
Fig.~\ref{fig:access_lowering_large}. We saw that the first and the
third dimensions of $X$ are cdims and correspond to the batch size
($s_1$) and the number of attention heads ($s_3$) respectively. The
other two dimensions, which correspond to sequence lengths are vdims,
the size of a slice for which is the same function ($s_{24}())$) of
the outermost batch dimension.

The sparse tensor compiler Taco~\cite{taco}, the performance of which
look at in~\S\ref{sec:ap_taco_eval}, uses a tree-based modular scheme
(first proposed in the work~\cite{csf} on the Compressed Sparse Fiber
tensor format) to model sparse tensor storage. In this scheme,
illustrated in Fig.~\ref{fig:access_lowering_large} for tensor $X$,
tensor storage is modeled as hierarchical tree structure, where each
tensor dimension corresponds to a tree level. Note that this tree
abstraction exists only at compile time. As mentioned before, this
scheme assumes that the number of non-zeros in a slice of a sparse
tensor dimension can depend on the indices of all outer dimensions in
general. We saw that this is not the case with ragged tensors and that
this is the source of sub-optimality in this lowering scheme for the
applications we look at. Because every slice may have a different
number of non-zero elements, when used to store a ragged tensor, this
storage scheme would store auxiliary data proportional to the number
of slices for a given vdim. For our example tensor $X$ in
Fig.~\ref{fig:access_lowering_large}, the outer of the two vdims (the
second dimension) has $s_1$ slices while the number of slices in the
inner vdim (the fourth dimension) is
$s_3\sum_{i=0}^{s_1}s_{24}(i)$. Therefore, the amount of auxiliary
data computed and stored would be equal to $s_1 +
s_3\sum_{i=0}^{s_1}s_{24}(i)$, which as we saw
in~\S\ref{sec:cora_oh_eval} can be much larger than \Sys's specialized
scheme.

\begin{algorithm}
  \vspace{-1mm}
  \caption{Procedure to lower ragged tensor accesses}
  \small
  \label{alg:lower_access}
  \begin{algorithmic}[1]
    \Procedure{LowerAccess}{$\lbrack b_1, ..., b_n \rbrack$}
    \State $\mathit{offset} \gets 0$
    \State $\mathit{relaxed} \gets [b_1, ..., b_n]$

    \For{$i \gets n$ to $1$} \Comment{Compute $D_i(\overrightarrow{B_{\le i}})$}    \label{line:outer_loop}
      \State $D \gets 1$
        \If{$O_G(i) \ne \emptyset$}
          \State $D \gets A_i(\mathit{relaxed}[j])$
        \Else
          \State $D \gets \mathit{relaxed}[i]$
        \EndIf

        \For{$j$ in $S(i) - \{i\}$}                                                     \label{line:inner_loop}
          \If{$O_G(j) \ne \emptyset$}                                                 \label{line:check}
            \State $D \gets D * A_j(\mathit{relaxed[j]})$
          \Else
            \State $D \gets D * s_j(\mathit{relaxed}[I_G(j)])$
          \EndIf
        \EndFor
      \State $\mathit{relaxed}[i] \gets s_i(\mathit{relaxed}[I_G(i)])$
      \State $\mathit{offset} \gets \mathit{offset} + D$
    \EndFor

    \State \Return $\mathit{offset}$
    \EndProcedure
  \end{algorithmic}
  \vspace{-1mm}
\end{algorithm}

\noindent\textbf{\Sys's Storage Access Lowering Scheme:} We saw that
\Sys's~storage access lowering scheme is specialized for ragged
tensors and enables us to reduce the amount of auxiliary data that
needs to be computed as compared to the scheme used by past work while
allowing $O(1)$ accesses to ragged tensor storage. Such $O(1)$
accesses are enabled by the memory offsets that \Sys~precomputes as
part of its auxiliary data structures. Below, we describe exactly how
these data structures are computed and how they are used to lower
memory accesses.

Let $T$ be an n-dimensional tensor with dimensions numbered $1$ to $n$
such that dimension $1$ is the outermost dimension. Given a tensor
access $T(b_1, .., b_n)$, we need to generate a flat memory access as
part of lowering. In other words, we need to generate a memory offset
$\mathit{Off}_T(b_1, .., b_n)$ to access the tensor.

Given a tensor and its corresponding storage layout, we define what we
refer to as the \emph{dimension graph} or \emph{dgraph} for short
(Fig.~\ref{fig:access_lowering_large}). The dgraph $G$ of the
n-dimensional tensor $T$ is a pair $(D, E)$ where $D$ is the set of
all dimensions $\{1, ..., n\}$ and $E$ is a set of directed edges. An
edge $d_1 \rightarrow d_2$ belongs to $E$ if the size of a slice of
dimension $d_2$ depends on the index $b_{d_1}$ in the tensor access
$T(b_1, .., b_n)$. Thus, a cdim will not have any incoming edge in the
dgraph, while a vdim would. It also follows, for example, that the
outermost dimension of the tensor, which is always a cdim, will not
have any incoming edges. More generally, we note that the dgraph of a
given tensor is always acyclic as the size of a slice of a given vdim
depends only on the indices of outer dimensions. Further, given a
dimension $d$, let $O_G(d) = \{d_2 | (d, d_2) \in E\}$ and $I_G(d) =
\{d_1 | (d_1, d) \in E\}$ be the set of outgoing and incoming
dimensions, respectively, for $d$ in the dimension graph. The size of
a slice of a vdim $d$ can now be written as $s_d(I_G(d))$. For cdims,
this quantity is constant as $I_G$ for a cdim in the empty set. Let
$O_G^{*}(d)$ denote the transitive closure of $O_G(d)$. Also, let
$O_G^{ex}(d) = O_G(d) - \bigcup_{i \in O_G(d)}O_G^{*}(i)$.

We present the procedure to compute $\mathit{Off}_T(b_1, .., b_n)$ in
Algorithm~\ref{alg:lower_access}. For brevity, we refer to the index
vector $[b_1, .., b_n]$ as $\overrightarrow{B}$. Also, let
$\overrightarrow{B_{\ge i}} = [b_i, ..., b_n]$. We can correspondingly
defined $\overrightarrow{B_{\le i}}$. We abuse notation to represent
$\mathit{Off}_T(b_1, ..., b_{i-1}, b_{i}, 0, ..., 0)$ as
$\mathit{Off}_T(\overrightarrow{B_{\le i}})$. Then, we can expand the
offset $\mathit{Off}_T(\overrightarrow{B_{\le n}})$ as follows:

\vspace{-5mm}
\begin{equation*}
  \begin{split}
    \mathit{Off}_T(\overrightarrow{B_{\le n}}) &= \sum_{i=1}^{n} (\mathit{Off}_T(\overrightarrow{B_{\le i}}) - \mathit{Off}_T(\overrightarrow{B_{< i}})) \\
                                               &= \sum_{i=1}^{n} D_i(\overrightarrow{B_{\le i}})
  \end{split}
\end{equation*}

During compilation, the procedure in Algorithm~\ref{alg:lower_access}
computes the memory offset expression using two nested loops. Each
iteration of the outer loop (line~\ref{line:outer_loop}) corresponds
to one dimension $i$ and computes $D_i(\overrightarrow{B_{\le
i}})$. For a dimension $i$, $D_i(\overrightarrow{B_{\le i}})$, is
further computed (in the inner loop on line~\ref{line:inner_loop}) as
a product of contributions corresponding each of the inner dimensions
$j$ such that $j \ge i$ (Fig.~\ref{fig:access_lowering_large} shows
the values of $D_i$s for the 4 dimensions in our example tensor $X$ at
the bottom of the tree in green in the rightmost pane.). In the case
of a dense tensor, $D_i(\overrightarrow{B_{\le i}}) =
b_i\prod_{j=i+1}^ns_j$. For a ragged tensor, however, due to the
dependencies between dimensions, the contribution of each dimension
$j$ to $D_i$ cannot be computed independently. Specifically, we
compute the contribution of an inner dimension $j$ along with all the
dimensions dependent on it, directly or indirectly (i.e. $O_G^{*}(j)$)
as a single quantity as a call to the function $A_j()$. This function
is similar to the \verb+row_index+ array in the CSR matrix format
which stores the start and ends of variable-sized rows. Given a ragged
tensor format (in the form of the length functions $s_d$ for all
dimensions $d$), we need to precompute the values of the function
$A_d$ for all dimensions such that $O_G(d)$ is non-empty. We perform
this computation as part of the prelude discussed
in~\S\ref{sec:overview}. The function $A_d()$ for the batch dimension
(the first dimension) of our example tensor $X$ in
Fig.~\ref{fig:access_lowering} is shown as the array \verb+A+$_1$
where
\verb+A+$_1$\verb+[+$i$\verb+]+$=\sum_{j=1}^{i}s_{24}(j)*s_{24}(j)$.

As discussed above, for a dimension $d$, because, $A_d()$ includes the
contributions from all dimensions in $O_G^{*}(d)$, we need to exclude
those dimensions to avoid double counting them during the inner
loop. Therefore, the inner loop of the procedure iterates over the set
$S(d)$ (defined recursively as $S(n) = \{n\}$ and $S(d) = \{d\} \cup
(S(d+1) - O_G^{*}(d))$) which excludes these dimensions. Given a
dimension $d$, the function $A_d$ can be computed recursively as
follows.

\vspace{-7mm}
\begin{equation*}
  \arraycolsep=1pt\def\arraystretch{2.2}
\resizebox{1.0\columnwidth}{!}{$%
A_d(B_{\le d}) = \left\{\begin{array}{lr}
                 s_d(B_{\le d}), & \text{if } O_G(d) = \emptyset\\
                 \sum\limits_{i = 0}^{i = b_d} (\prod\limits_{d_i \in O_G^{ex}(d_i)} A_{d_i}(relaxed_d[I_G(d_i)])) & \text{otherwise}
\end{array}\right.
$}
\end{equation*}

where $relaxed_d$ is the value of the vector $relaxed$ in
Algorithm~\ref{alg:lower_access} in the iteration of the outer loop
corresponding to the dimension $d$.

\subsection{Variable Loop Fusion}\label{sec:ap_vloop_fuse}
In~\S\ref{sec:fusion}, we discussed how we need to precompute certain
quantities as part of the prelude to support vloop fusion. During
lowering, we represent these quantities as opaque or uninterpreted
functions. For example, \S\ref{sec:bounds_inference} describes how the
functions $f_{fo}$, $f_{fi}$ and $f_{oif}$ represent the relationships
between the iteration variables \verb+o+, \verb+i+ and \verb+f+ in
Fig.~\ref{fig:fusion}. In the generated code, as we can see in
Fig.~\ref{fig:overview}, these functions take the form of arrays that
are initialized by the prelude. During compilation, in order to
perform simplification over expressions containing calls to these
functions as well as for proving if certain bound checks are
redundant, we use the Z3 SMT solver~\cite{z3}. In order to enable Z3
process these uninterpreted functions, we provide it with the
following relationships between these functions:

\vspace{-5mm}
\begin{align*}
  \forall f, f_{oif}(f_{fo}(f), f_{fi}(f)) &= f \\
  \forall o, i, f_{fo}(f_{oif}(o, i)) &= o \\
  \forall o, i, f_{fi}(f_{oif}(o, i)) &= i
\end{align*}

\section{Additional Implementation Details} \label{sec:ap_impl}
As we mentioned in~\S\ref{sec:impl}, we have prototyped \Sys~for the
common cases encountered when expressing and optimizing ragged
operations. In our evaluation, we implement and compare the
performance of an encoder layer of the transformer model in \Sys. Our
prototype currently allows us to generate code for individual
(potentially fused) ragged operators at a time as opposed to entire
model graphs. Therefore, for our implementation of the transformer
layer, we individually optimized and generated code for each operator
and then invoked it as part of a separate program that ties the
operators together to form the layer. \Sys's implementation of the
hfusion optimization currently is limited to the outermost loops of
the operators one would like to fuse. On a GPU, this means that our
prototype implementation allows one to execute multiple operators
concurrently as part of the same GPU grid, but not the same GPU thread
block. Implementing the general transform is not fundamentally
difficult, however.

\section{Supplementary Evaluation and Additional Details} \label{sec:ap_eval}
\subsection{Datasets} \label{sec:ap_datasets}
We use the datasets listed in Table~\ref{table:datasets} for the
evaluation on the transformer model. For each dataset, we use the
sequence lengths corresponding to the text obtained after
preprocessing as performed in the implementations corresponding to
past work on various transformer models~\cite{transformer, bert,
xlnet}. The Wiki512 and Wiki128 datasets, usually used for
pre-training~\cite{bert}, are generated from a dump of the English
Wikipedia website~\cite{wikipedia}. Each sequence in these two
datasets was created by accumulating consecutive sentences from the
dump until a sentence could no longer be added without exceeding the
maximum sequence length used for training (which is a
hyperparameter). This was done, in the transformer implementation, to
reduce wasted computation due to padding as much as possible. As a
result, these datasets do not provide as much opportunity for \Sys~to
exploit as do some of the other datasets. We saw this reflected in
Fig.~\ref{fig:flop_ratios} as well as in the evaluation
in~\S\ref{sec:eval}.

\subsection{Load Balancing}
We briefly discussed the challenge of ensuring a balanced workload
across multiple execution units in the main text. On a CPU, these
execution units take the form of CPU cores, while a GPU has a
hierarchy composed of thread blocks, warps and threads. In all the
kernels we evaluate on (except the Softmax kernel in the transformer
layer), dense inner cloops or partial padding allow us to prevent
imbalance across GPU warps in the same thread block. Imbalance across
multiple thread blocks exists, most commonly in gemm-like operations
where the reduction loop is a vloop such as the AttnV operator in the
SDPA module. We handle this imbalance using either thread remapping
(\S\ref{sec:api} and~\S\ref{sec:ap_remap}) or, in the case of kernels
that are part of the transformer layer, by sorting the sequences in
the mini-batch in descending order of sequence lengths so that thread
blocks with the most amount of work are scheduled first.

\subsection{Masked Scaled Dot-Product Attention} \label{sec:ap_mmha_eval}
As we briefly mentioned in~\S\ref{sec:layer_eval}, the decoder layer
of a transformer uses a variant of MHA called masked MHA wherein the
upper triangular half of the attention matrix is masked for all
attention heads during training. This is done to prevent the model
from attending to words that would not be known during inference at a
given time step.  In this section, we provide further details and data
regarding how \Sys~can exploit this masking and further save on wasted
computation in the SDPA sub-module, which is the only portion affected
by the masking.

\begin{figure}[htpb]
  \centering
  \includegraphics[width=0.70\columnwidth]{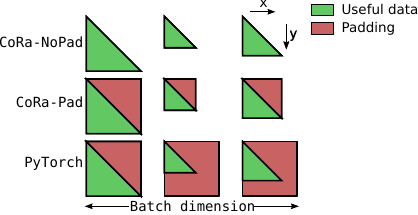}
  \caption{The attention matrices of the masked MHA module as
    implemented in the implementations discussed
    in~\S\ref{sec:layer_eval} and compared in
    Fig.~\ref{fig:ap_mmha_eval}. In the figure, for simplicity, the
    number of attention heads is assumed to be 1, partial padding is
    not shown and the batch size is assumed to be 3. The \texttt{x}
    and \texttt{y} directions denote increasing matrix indices.}
  \label{fig:ap_mmha_fig}
\end{figure}

\begin{figure}[htpb]
  \centering \includegraphics[width=0.90\columnwidth]{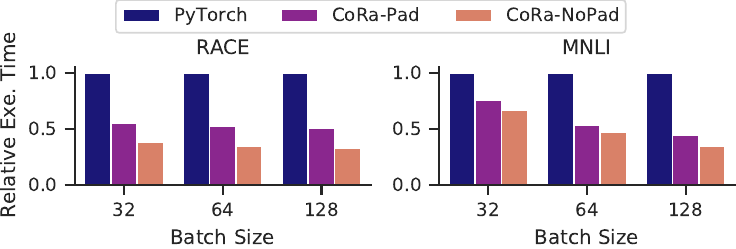}
  \caption{Execution time of masked SDPA in PyTorch and \Sys, with and
    without padding for the attention matrix.}
  \label{fig:ap_mmha_eval}
\end{figure}

We also mentioned in the main text that with masking, the SDPA
computation is essentially composed of batched lower triangular matrix
operations. Implemented this way, these operations have one vloop
corresponding to the variable sequence lengths and another inner vloop
corresponding to the triangular matrix
rows. Fig.~\ref{fig:ap_mmha_eval} shows the performance of three
implementations of masked SDPA---\Sys-NoPad, where both the vloops are
only partially padded, \Sys-Pad, where the outer vloop is partially
padded while the inner one is fully padded and a PyTorch
implementation, where both the vloops are fully padded. The padding in
the three implementations is illustrated in
Fig.~\ref{fig:ap_mmha_fig}. As Fig.~\ref{fig:ap_mmha_eval} shows,
\Sys-NoPad can effectively exploit the reduction in computation in the
masked case by avoiding full padding. This leads to $1.34\times$ and
$2.46\times$ faster execution as compared to \Sys-Pad and PyTorch
respectively across the datasets and batch sizes evaluated in
Fig.~\ref{fig:ap_mmha_eval}. As we saw, the performance of MNLI
dataset improves to a smaller degree due to the padding employed in
\Sys-NoPad.

\subsection{Evaluation Against Sparse Tensor Compilers}\label{sec:ap_taco_eval}
We saw in the main text of the paper that there are some similarities
between ragged and sparse tensors. In this section, we explore using
sparse tensor compilers in order to express ragged tensor
operations. Specifically, we look at using Taco in order to implement
three operations on triangular matrices---the triangular matrix
multiplication (trmm) operation we saw in~\S\ref{sec:vtrmm_eval},
elementwise addition of two square triangular matrices (we refer to
this operation as tradd, for short) and a similar elementwise
multiplication of two square triangular matrices (referred to as
trmul, for short). Taco does not natively support the storage of
ragged tensors. Therefore for this study, we use the compressed sparse
row (CSR) and the blocked compressed sparse row (BCSR) matrix formats
to store the triangular matrices. We use a block size of 32 for the
BCSR format. Table~\ref{table:taco_eval} lists the execution times (in
ms) for the aforementioned operations and formats. As the table shows,
\Sys~performs better than Taco for all the cases evaluated. We discuss
the reasons for this below.

\noindent\textbf{Storage Layouts:} A part of the slowdown in Taco
stems from the sub-optimal storage format (CSR or BCSR) used for the
triangular matrices. The overheads of traversing the auxiliary data
structures to access the sparse tensor storage therefore decrease when
we go from the CSR format to the BCSR format, thereby leading to
increased performance, despite the additional padding in the
latter. For the operations evaluated, the output matrices are stored
in a dense manner because using the compressed formats prevents
parallelization in some cases in the Taco implementations.

\begin{table}
  \centering
  \scriptsize
  \caption{Execution times (in ms) for the trmm, tradd and trmul
    operations implemented in Taco using the CSR and the BCSR matrix
    formats and in \Sys. The table also shows Taco's slowdowns with
    respect to \Sys.}
  \resizebox{0.98\columnwidth}{!}{%
  \addtolength{\tabcolsep}{-3pt}
  \begin{tabular}{ccccccc}
    \toprule
    \multirow{2}{*}{\centering Op} & \multirow{2}{*}{\centering Matrix Dim.} & \multirow{2}{*}{\centering \Sys} & \multicolumn{2}{c}{\centering Taco-CSR} & \multicolumn{2}{c}{\centering Taco-BCSR} \\
    \cmidrule(lr){4-7}
       &             &      & Time & Slowdown                & Time & Slowdown                 \\ \midrule
    \multirow{4}{*}{trmm} & 128 & 0.043 & 0.062 & 1.44 & 0.467 & 10.92 \\
 & 512 & 0.082 & 1.347 & 16.43 & 1.112 & 13.56 \\
 & 2048 & 0.893 & 75.12 & 84.19 & 47.497 & 53.24 \\
 & 8192 & 50.905 & 4854.31 & 95.37 & 4252.33 & 83.54 \\ \hline
\multirow{4}{*}{tradd} & 128 & 0.004 & 0.057 & 15.61 & - & - \\
 & 512 & 0.004 & 0.223 & 61.68 & - & - \\
 & 2048 & 0.033 & 1.538 & 46.94 & - & - \\
 & 8192 & 0.476 & 7.883 & 16.58 & - & - \\ \hline
\multirow{4}{*}{trmul} & 128 & 0.004 & 0.057 & 15.89 & 0.008 & 2.08 \\
 & 512 & 0.004 & 0.225 & 57.21 & 0.016 & 3.87 \\
 & 2048 & 0.033 & 1.544 & 47.26 & 0.077 & 2.34 \\
 & 8192 & 0.476 & 7.92 & 16.67 & 0.632 & 1.33 \\ \bottomrule

  \end{tabular}
  \addtolength{\tabcolsep}{3pt}
}
  \label{table:taco_eval}
\end{table}

\noindent\textbf{Degree of Sparsity:} The optimizations, scheduling
primitives and code generation techniques used in Taco have been
designed for tensors with a high degree of sparsity. We have seen,
however, that ragged tensors are much closer to their dense
counterparts with respect to the amount of useful data they
store. Therefore, optimization decisions that work well for sparse
tensors do not always work for ragged tensors.

\noindent\textbf{Properties of Ragged Tensors:} Finally, due to its
design as a tensor compiler for general sparse tensors, Taco is unable
to exploit certain properties specific to ragged tensors and the
applications they are used for, such as the
insight~\ref{prop:known_raggedness} we discussed
in~\S\ref{sec:overview}. Therefore, Taco assumes that the two
triangular input matrices in the tradd and trmul operations have
differing sparsity patterns. Taco, therefore, has to generate code to
iterate over all the coordinates representing the union of the
non-zeroes in the input matrices for the tradd operator. This is
unlike an intersection that is performed in trmul. This prevented us
from scheduling the tradd operator using the BCSR format in a way
similar to the trmul operator. Further, Taco currently does not allow
users to specify padding for loops and tensor dimensions which would
help elide conditional checks in the generated code.

Therefore, while Taco achieves performance comparable to \Sys's in some
cases (such as the trmul operator), we conclude that Taco's
programming model and optimizations are designed for highly sparse
tensors which can lead to poor performance in a lot of cases involving
ragged tensors.

\subsection{Memory Consumption} \label{sec:ap_mem}
\begin{figure}[htpb]
  \centering
  \includegraphics[width=0.90\columnwidth]{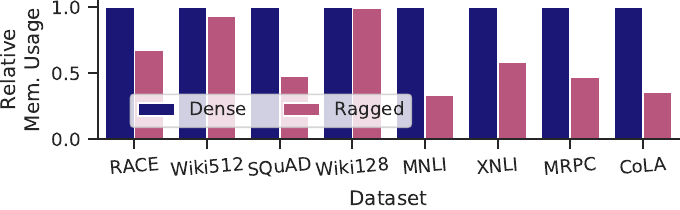}
  \caption{Relative sizes of the forward activations of a transformer
    encoder layer with and without ragged tensors.}
  \label{fig:mem_eval}
\end{figure}

We mentioned in~\S\ref{sec:layer_eval} that the use of ragged tensors
leads to a significant drop in the memory required to store the
forward activations of the encoder layer. Fig.~\ref{fig:mem_eval}
shows this for the datasets in Table~\ref{table:datasets} for batch
size 64. It plots the relative total memory consumption (computed
analytically) of the forward activations of a transformer encoder
layer for \Sys's implementation with and without the use of ragged
tensors. We take into account any partial padding that the ragged
implementation requires. The relative memory consumption for the other
batch sizes is also similar. We also saw how only small improvements
are observed for the Wiki512 and Wiki128 datasets which have higher
sequence lengths and hence low opportunity for \Sys~to exploit.

\subsection{Operation Splitting and Horizontal Fusion} \label{sec:ap_bp}

\begin{figure}[htpb]
  \centering
  \includegraphics[width=0.90\columnwidth]{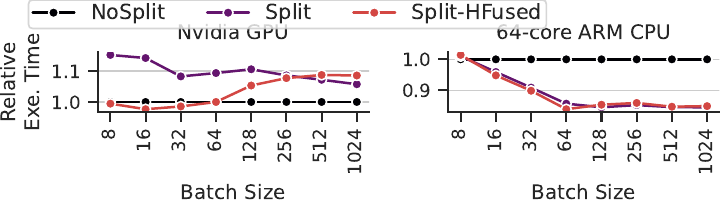}
  \caption{Operation splitting and hfusion for QK\textsuperscript{T}.}
  \label{fig:ap_qkt_bp_eval}
\end{figure}

In~\S\ref{sec:bp_eval} of the main text, we looked at the benefits of
operation splitting and hfusion on the AttnV operator. We now look at
the QK\textsuperscript{T} operator, which is also an instance of the
vgemm problem. Each gemm instance in this case has two non-reduction
vloops. We first look at the case where the optimizations are applied
to the outer one of these two vloops in
Fig.~\ref{fig:ap_qkt_bp_eval}. The figure shows the normalized
execution times, for the QK\textsuperscript{T} operator, of the three
implementations described in~\S\ref{sec:bp_eval}. We see that on the
CPU backend, similar to the AttnV operator, operation splitting has a
significant benefit but hfusion does not, due to low parallelism
exposed by the CPU. On the GPU backend, however, we see that the
combination of the optimizations gives slightly better performance for
lower batch sizes but performs worse as the batch size
increases. Profiling data shows that applying the optimization in this
case leads to an increase in the number of integer instructions
executed as well as an increase in the number of memory load
requests. One possible explanation for this is that the CUDA compiler
does not effectively hoist memory access expressions in order to avoid
high register pressure (the compiled code does not contain any spilled
registers). While the optimizations generally lead to more complicated
code, the fact that QK\textsuperscript{T} has two vloops that we fuse
when scheduling further exacerbates this problem.

\begin{figure}[htpb]
  \centering
  \includegraphics[width=0.90\columnwidth]{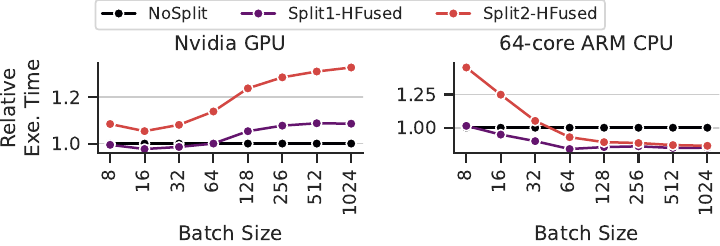}
  \caption{Efficacy of operation splitting and hfusion when applied to
    one or both vloops of the QK\textsuperscript{T} operator.}
  \label{fig:ap_qkt_bp_split_eval}
\end{figure}

When applied to both the vloops, the optimizations slow the execution
down as seen in Fig.~\ref{fig:ap_qkt_bp_split_eval}. In that figure,
we compare the performance of three
\Sys~implementations---NoSplit, which does not use either of the
optimizations on either vloop, Split1-HFused, which employs both the
optimizations for the outer vloop and Split2-HFused, which employs the
optimizations for both vloops---on the Nvidia GPU and the 64-core ARM
CPU backends. We see that on both backends, optimizing both vloops is
no better than optimizing just one vloop and is, in fact, quite slower
on the GPU. On the GPU, we find that despite the decrease in the
computation performed and hence the number of floating point
instructions executed, the total number of executed instructions is
higher in the case Split2-HFused case as compared to the NoSplit
case. We therefore believe, that in this case too, the overheads of
performing the optimizations are much higher than their benefits (the
reduced wasted computation).

\subsection{\Sys~Overheads} \label{sec:ap_cora_overheads}

\begin{table*}[htpb!]
  \centering
  \scriptsize
  \caption{Prelude execution times (in ms) for a 6-layer transformer
    encoder with and without redundant computation.}
\begin{tabular}{cc | ccc | ccc}
  \toprule
  \multirow{2}{*}{\centering Dataset} &
  \multirow{2}{*}{\centering Batch Size} &
  \multicolumn{3}{c|}{\Sys-Redundant}  &
  \multicolumn{3}{c}{\Sys-Optimized} \\

  &
  &
  \Sys~Storage & \Sys~Loop Fusion & \Sys-Copy Time &
  \Sys~Storage & \Sys~Loop Fusion & \Sys-Copy Time \\ \midrule
  CoLA & 32 & 0.004 & 0.006 & 0.232 & 0.002 & 0.002 & 0.088 \\
CoLA & 128 & 0.006 & 0.015 & 0.261 & 0.003 & 0.004 & 0.094 \\
RACE & 32 & 0.005 & 0.085 & 0.419 & 0.002 & 0.015 & 0.121 \\
RACE & 128 & 0.007 & 0.339 & 0.985 & 0.003 & 0.053 & 0.209 \\ \bottomrule

\end{tabular}
\label{table:ap_prelude_time}
\end{table*}

\noindent\textbf{Prelude Overheads:} As we discussed
in~\S\ref{sec:ap_impl}, \Sys's prototype allows us to generate code
for operator kernels one at a time. For each kernel, \Sys~generates
all the prelude code required for its execution. Therefore, when these
generated kernels are invoked to form a larger model graph, as in our
implementation of the transformer encoder layer, there is a lot of
redundant computation in the prelude code. This is because (i) each
operator computes the auxiliary data structures needed for all of its
input and output tensors, which leads to these data structures being
generated twice for every tensor in the graph, and (ii) the vloops in
the schedules for all operators except the QK\textsuperscript{T} and
the AttnV operators in \Sys's implementation of the layer are fused
similarly and can reuse the same auxiliary data structures, which are
also currently computed separately for every
operator. Tables~\ref{table:ap_prelude_time}
and~\ref{table:ap_prelude_mem} compare, for a 6-layer transformer
encoder, the execution time and memory consumption of the prelude code
respectively, as present in \Sys's current implementation (referred to
as \Sys-Redundant in the table) with an optimized implementation
(referred to as \Sys-Optimized) which has all of this redundant
computation removed. We see that when appropriately reused, the time
and memory resources required to compute the auxiliary data structures
in the prelude are quite low as compared to the those required for the
execution of the kernel computation.

\begin{table}[htbp!]
  \centering
  \scriptsize
  \caption{Prelude memory usage (in kB) for a 6-layer transformer
    encoder with and without redundant computation.}
  \addtolength{\tabcolsep}{-2pt}
  \begin{tabular}{p{1cm} p{1cm} | p{1.0cm} p{1.3cm} | p{1.0cm} p{1.3cm}}
  \toprule
  \multirow{3}{1cm}{Dataset} &
  \multirow{3}{1cm}{Batch Size} &
  \multicolumn{2}{p{2.3cm} | }{\centering \Sys-Redundant}  &
  \multicolumn{2}{p{2.3cm}}{\centering \Sys-Optimized} \\

  &
  &
  {\centering \Sys\ Storage} & {\centering \Sys\ Loop Fusion} &
  {\centering \Sys\ Storage} & {\centering \Sys\ Loop Fusion} \\ \midrule
  CoLA & 32 & 2.93 & 32.15 & 1.2 & 5.27 \\
CoLA & 128 & 11.18 & 104.22 & 4.58 & 17.5 \\
RACE & 32 & 2.93 & 666.54 & 1.2 & 106.87 \\
RACE & 128 & 11.18 & 2609.58 & 4.58 & 418.06 \\ \bottomrule

  \end{tabular}
  \addtolength{\tabcolsep}{2pt}
  \label{table:ap_prelude_mem}
\end{table}

\begin{figure}[htpb]
  \centering
  \includegraphics[width=0.90\columnwidth]{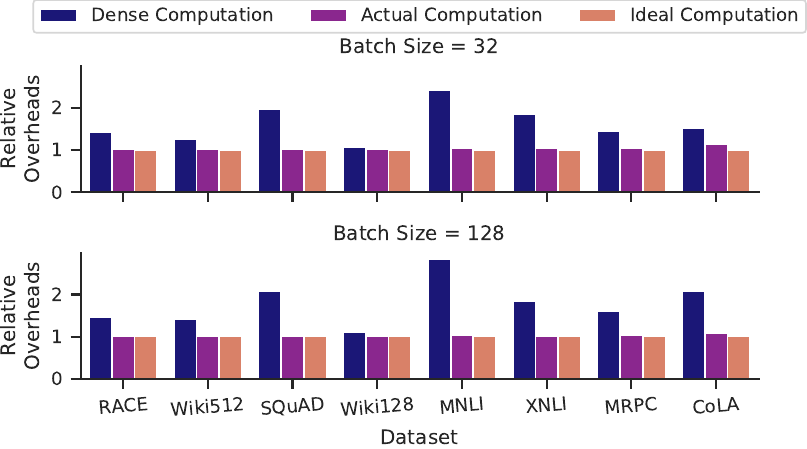}
  \caption{Overheads due to partial padding. }
  \label{fig:partial_padding_oh_eval}
\end{figure}

\noindent\textbf{Overheads Due to Partial Padding:} In
Fig.~\ref{fig:partial_padding_oh_eval}, we show the relative amount of
computation (computed analytically as in Fig.~\ref{fig:flop_ratios})
for the transformer encoder layer for all datasets at batch sizes 32
and 128 for three cases---the fully padded dense case, the actual
computation as evaluated in~\S\ref{sec:eval} with partial padding, and
the ideal case with no padding. We see that partial padding leads to a
very small increase in the amount of computation (3.5\% across
datasets for batch size 32 and 2.3\% for batch size 128). Because we
generally pad individual sequence lengths or their sum (as part of
bulk padding) so that the quantity is a constant multiple of a small
quantity (such as 32, or 64), the relative amount of padding added is
higher for smaller batch sizes and datasets with smaller sequence
lengths. Even in these cases, however, the added padding is much lower
as compared to the benefits obtained with the use of ragged
tensors. Further we note that the amount of padding added is a
scheduling and optimization decision and can be changed if needed.

\begin{figure}[htpb]
  \centering
  \includegraphics[width=0.90\columnwidth]{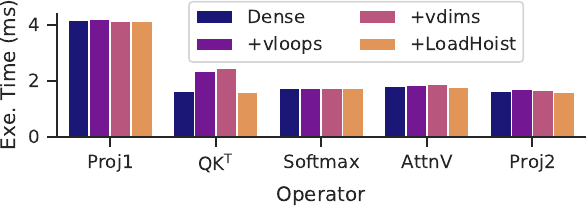}
  \caption{Overheads of using ragged computations and ragged tensor
    storage, and the benefits of load hoisting, measured for a
    synthetic dataset where all sequence lengths are 512. The batch
    size used is 64.}
  \label{fig:ragged_oh_eval}
\end{figure}

\noindent\textbf{Ragged Tensor Overheads and Load Hoisting:} We now
take a closer look at the effects of auxiliary data structure accesses
on the performance of \Sys-generated code. These data structure
accesses arise in the generated code, as we have seen, due to the use
of vloop fusion and ragged tensor storage. We focus on the five
operators that make up the MHA module here. We measure the execution
times of four implementations of each operator. The Dense
implementation does not use ragged tensor storage or ragged
computations. The +vloops implementation uses ragged computations, but
the tensors are stored with full padding in a dense fashion. The
+vdims implementation uses both ragged computations as well as ragged
tensor storage. The +LoadHoist implementation is same as +vdims but
hoists accesses to the auxiliary data structures out of loops as much
as possible. In order to ensure that we perform the same amount of
computation in all cases, we use a synthetic dataset where all
sequences have the same length (512). The relative performance of
these implementations for the operators on the Nvidia GPU is shown in
Fig.~\ref{fig:ragged_oh_eval}. Apart from the overheads due to
indirect memory accesses, the use of vloops and/or vdims also lead to
overheads associated with the prelude code. In order to focus on the
former overheads, however, we exclude prelude costs in the figure.

As the figure shows, the use of vloops and vdims leads to a slight
slowdown for the Proj1, Softmax, Attnv and Proj2 operators. The
slowdown is significant, however, for the QK\textsuperscript{T}
operator, which has two vloops in its loop nest. As part of
scheduling, we fuse both these vloops as well as the loop that the
vloop bounds depend on (i.e. the loop that iterates over the
mini-batch), leading to complex auxiliary data structure accesses. We
believe that the CUDA compiler is unable to effectively hoist these
accesses in this case. \Sys~however has more knowledge about these
accesses and can hoist them to recover the lost performance.

\subsection{Discussion on Transformer Layer Evaluation} \label{sec:ap_breakdown}
In this section, we provide further analysis of our evaluation of the
transformer encoder layer on the Nvidia GPU and ARM CPU backends. We
break down the execution time of the encoder layer for a few cases. As
in Fig.~\ref{fig:per_ops_plot}, these per-operator execution times are
obtained under profiling and might deviate slightly from the data in
Tables~\ref{table:layer_eval} and~\ref{table:arm_layer_eval}.

\noindent\textbf{Nvidia GPU Backend:} Table~\ref{table:ap_op_times}
provides the raw data for the breakdown of the execution times for the
RACE dataset at batch size 128 of the transformer encoder layer shown
in Fig.~\ref{fig:per_ops_plot} in the main text. Apart from
improvements in the QK\textsuperscript{T} and AttnV operators
discussed in~\S\ref{sec:layer_eval}, we note that \Sys's
implementation is significantly faster for the Softmax operator as
compared to the FasterTransformer implementations. While we perform
less computation on this operator as compared to the fully padded
implementation in FasterTransformer, part of \Sys's performance
benefits also stem from a better schedule. Specifically, the
FasterTransformer implementation performs parallel reductions across
GPU thread blocks. This leads to a significant number of barriers at
the thread block-level which have execution overheads. Further, the
FasterTransformer implementation uses conditional checks to ensure
that it never accesses attention scores for the added padding. In
\Sys~we use warp-wide parallel reductions which are much cheaper due
to their lower synchronization costs but also provide a lower amount
of parallelism. We, therefore, only partially parallelize the
reductions and compensate with the high parallelism available in the
other loops of the operator. Further, this means that we do not have
to additionally employ conditional checks to avoid accessing invalid
data (that is part of the partial padding we add).

\begin{figure}[htpb]
  \centering
  \includegraphics[width=0.90\columnwidth]{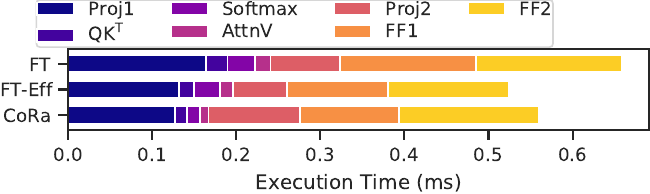}
  \caption{Breakdown of execution times of the encoder layer for the
    CoLA dataset at batch size 32 on the GPU.}
  \label{fig:ap_per_ops_plot_cuda_cola}
\end{figure}

\begin{figure}[htpb]
  \centering
  \subcaptionbox{MNLI dataset at batch size
    128.\label{fig:ap_per_ops_plot_arm_wiki}}{
    \includegraphics[width=0.9\columnwidth]{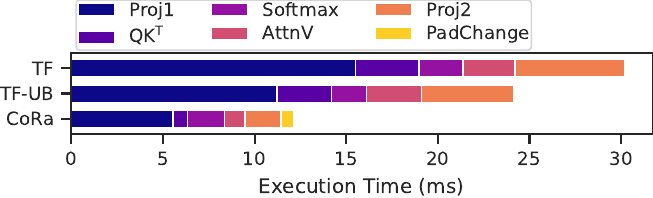}
  }%
  \vspace{2mm}
  \subcaptionbox{Wiki128 dataset at batch size
    32.\label{fig:ap_per_ops_plot_arm_wiki}}{
    \includegraphics[width=0.9\columnwidth]{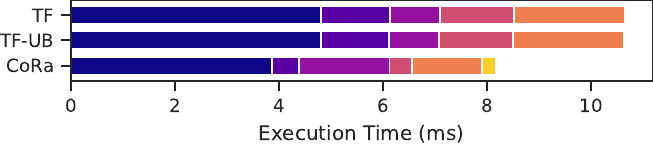}
  }%
  \vspace{2mm}
  \subcaptionbox{CoLA dataset at batch size
  32.\label{fig:ap_per_ops_plot_arm_race}} {
    \includegraphics[width=0.9\columnwidth]{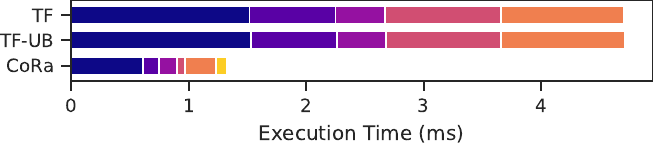}
  }%
  \vspace{2mm}
  \subcaptionbox{RACE dataset at batch size 128.\label{fig:ap_per_ops_plot_arm_race}} {
    \includegraphics[width=0.9\columnwidth]{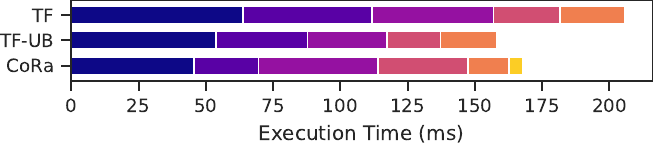}
  }%
  \vspace{-2mm}
  \caption{Breakdown of execution times of the MHA module for four
    cases on the 64-core ARM CPU
    backend.}
  \label{fig:ap_per_ops_plot_arm}
\end{figure}

We now look at the execution time breakdown for the CoLA dataset at
batch size 32 on the Nvidia GPU shown in
Fig.~\ref{fig:ap_per_ops_plot_cuda_cola}. We see that \Sys~performs
slightly worse than FT-Eff for this case. Most of \Sys's slowdown
stems from worse performance on the linear transformation operators
Proj2, FF1 and FF2. \Sys~performs slightly better than FT-Eff for the
Proj1 operator, which is also a linear transformation operator. From
this data, we conclude that \Sys's schedules for the Proj2, FF1 and
FF2 operators can be improved to close this performance gap. We note
that, even in this case, \Sys~performs much better on the SDPA module
(the QK\textsuperscript{T}, Softmax and AttnV operators) as compared
to FasterTransformer.

%% \begin{table*}[htpb!]
  %% \vspace{-11cm}
  %% \centering
  %% \scriptsize
  %% \caption{Prelude memory consumption (in kB) for a 6-layer
    %% transformer encoder with and without redundant computation.}
%% \begin{tabular}{cc | cc | cc}
  %% \toprule
  %% \multirow{2}{1cm}{\centering Dataset} &
  %% \multirow{2}{1cm}{\centering Batch Size} &
  %% \multicolumn{2}{c|}{\Sys-Redundant}  &
  %% \multicolumn{2}{c}{\Sys-Optimized} \\

  %% &
  %% &
  %% \Sys{} Storage & \Sys{} Loop Fusion &
  %% \Sys{} Storage & \Sys{} Loop Fusion \\ \midrule
  %% \input{src/lowering_oh_table_appendix_mem.tex}
%% \end{tabular}
  %% \label{table:ap_prelude_mem}
%% \end{table*}

\noindent\textbf{ARM CPU Backends:} In~\S\ref{sec:layer_eval}, we saw
how \Sys~performs better than TensorFlow for the MHA module on the 8-
and 64-core ARM CPUs. In this section, we discuss these
implementations in more detail and provide more extensive evaluation.

%%%%%%%%%%%%%%%%%%%%%%%%%%%%%%%%%%%%%%%%%
\begin{table*}[htpb!]
  \centering \scriptsize \vspace{-3.5mm} \caption{MHA execution
  latencies (in ms) on 8- and 64-core ARM CPUs. uBS stands for the
  optimal micro-batch size.}

    \addtolength{\tabcolsep}{-2pt}
    \begin{tabular}{
        cc |
        ccccc |
        ccccc
      }
      \toprule
      \multirow{2}{*}{\centering Dataset} &
      \multirow{2}{*}{\centering Batch Size} &
      \multicolumn{5}{c|}{8-core ARM CPU} &
      \multicolumn{5}{c}{64-core ARM CPU} \\ \cmidrule(lr){3-12}

      & & PT & PT-UB / uBS & TF & TF-UB / uBS & \Sys & PT & PT-UB / uBS & TF & TF-UB / uBS & \Sys \\ \midrule
    \multirow{3}{*}{RACE} & 32 & 627 & \textbf{209} / 2 & 300 & 228 / 8 & 263 & 4373 & 127 / 2 & 55 & 46 / 16 & \textbf{44} \\
 & 64 & 1267 & \textbf{411} / 2 & 596 & 432 / 8 & 515 & 8724 & 253 / 2 & 111 & 88 / 32 & \textbf{85} \\
 & 128 & 2558 & \textbf{810} / 2 & 1189 & 835 / 8 & 1009 & 17431 & 511 / 2 & 209 & \textbf{156} / 32 & 168 \\ \hline
\multirow{3}{*}{Wiki512} & 32 & 620 & \textbf{227} / 2 & 294 & 246 / 8 & 285 & 4294 & 123 / 2 & 53 & 53 / 32 & \textbf{47} \\
 & 64 & 1267 & \textbf{443} / 2 & 597 & 466 / 8 & 561 & 8727 & 239 / 2 & 106 & 96 / 32 & \textbf{91} \\
 & 128 & 2563 & \textbf{875} / 2 & 1184 & 904 / 16 & 1094 & 17427 & 660 / 2 & 205 & \textbf{172} / 32 & 176 \\ \hline
\multirow{3}{*}{SQuAD} & 32 & 324 & \textbf{101} / 4 & 189 & 117 / 8 & 113 & 1904 & 94 / 4 & 35 & 27 / 16 & \textbf{20} \\
 & 64 & 770 & \textbf{192} / 4 & 383 & 210 / 8 & 219 & 4953 & 181 / 4 & 68 & 49 / 32 & \textbf{39} \\
 & 128 & 1580 & \textbf{364} / 4 & 780 & 390 / 8 & 424 & 10236 & 357 / 4 & 137 & 79 / 32 & \textbf{76} \\ \hline
\multirow{3}{*}{Wiki128} & 32 & 53 & \textbf{52} / 16 & 53 & 52 / 32 & 54 & 76 & 76 / 32 & 11 & 11 / 32 & \textbf{9} \\
 & 64 & 133 & 101 / 16 & 101 & \textbf{100} / 64 & 102 & 330 & 141 / 16 & 19 & 18 / 64 & \textbf{17} \\
 & 128 & 353 & 196 / 16 & 199 & \textbf{190} / 64 & 200 & 1544 & 273 / 16 & 34 & 33 / 128 & \textbf{33} \\ \hline
\multirow{3}{*}{MNLI} & 32 & 41 & 26 / 8 & 39 & 29 / 8 & \textbf{20} & 69 & 30 / 4 & 9 & 9 / 32 & \textbf{4} \\
 & 64 & 100 & 47 / 8 & 82 & 52 / 16 & \textbf{38} & 204 & 51 / 8 & 16 & 14 / 32 & \textbf{7} \\
 & 128 & 260 & 90 / 16 & 177 & 93 / 16 & \textbf{76} & 399 & 87 / 16 & 30 & 23 / 64 & \textbf{14} \\ \hline
\multirow{3}{*}{XNLI} & 32 & 53 & 36 / 8 & 52 & 42 / 16 & \textbf{33} & 76 & 58 / 2 & 11 & 11 / 32 & \textbf{6} \\
 & 64 & 133 & 68 / 8 & 101 & 73 / 16 & \textbf{65} & 324 & 95 / 8 & 18 & 18 / 64 & \textbf{11} \\
 & 128 & 351 & 131 / 16 & 199 & 134 / 32 & \textbf{128} & 1549 & 179 / 16 & 34 & 28 / 64 & \textbf{22} \\ \hline
\multirow{3}{*}{MRPC} & 32 & 38 & 31 / 8 & 37 & 33 / 16 & \textbf{27} & 71 & 46 / 4 & 9 & 8 / 32 & \textbf{5} \\
 & 64 & 86 & 59 / 8 & 75 & 61 / 16 & \textbf{52} & 172 & 80 / 8 & 14 & 14 / 64 & \textbf{10} \\
 & 128 & 187 & 110 / 16 & 151 & 111 / 32 & \textbf{103} & 351 & 153 / 8 & 26 & 23 / 64 & \textbf{18} \\ \hline
\multirow{3}{*}{CoLA} & 32 & 10 & 9 / 16 & 12 & 11 / 32 & \textbf{8} & 7 & 7 / 16 & 5 & 4 / 32 & \textbf{2} \\
 & 64 & 21 & 16 / 16 & 21 & 18 / 32 & \textbf{14} & 11 & 13 / 16 & 6 & 6 / 64 & \textbf{3} \\
 & 128 & 46 & 29 / 32 & 37 & 29 / 32 & \textbf{25} & 23 & 18 / 32 & 9 & 8 / 128 & \textbf{5} \\ \bottomrule

    \end{tabular}
    \addtolength{\tabcolsep}{2pt}
  \label{table:arm_fuller_eval}
\end{table*}
%%%%%%%%%%%%%%%%%%%%%%%%%%%%%%%%%%%%%%%%%

\begin{figure}
  \centering
  \includegraphics[width=0.95\columnwidth]{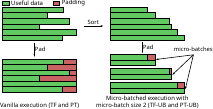}
  \caption{Comparison of vanilla and micro-batched execution for
  PyTorch and TensorFlow.}
\label{fig:micro_batch_fig}
\end{figure}

%%%%%%%%%%%%%%%%%%%%%%%%%%%%%%%%%%%%%%%%%
\noindent\underline{Micro-Batching for PyTorch and TensorFlow:}
We saw, in Fig~\ref{fig:flop_ratios}, that the amount of padding and
wasted computation increases with the batch size. On devices that
expose low levels of parallelism such as CPUs, it is therefore
possible to trade-off batch parallelism for reduced padding, and
therefore reduced wasted computation, for frameworks such as PyTorch
and TensorFlow. In effect, this amounts to executing a mini-batch
sorted by sequence lengths as a series of
smaller \emph{micro-batches}. Overall, this reduces the amount of
padding needed as each micro-batch is only padded to the length of the
longest sequence in that micro-batch, rather than the entire
mini-batch as illustrated in Fig.~\ref{fig:micro_batch_fig}. We search
over micro-batch sizes that are powers of 2 starting from the lowest
micro-batch size of 2. In Table~\ref{table:arm_fuller_eval}, we
provide the execution latencies as well as the optimal micro-batch
sizes for PyTorch and TensorFlow (these configurations is referred to
as PT-UB and TF-UB respectively) for an 8-core as well as a 64-core
ARM CPU. For reference, we also provide the latencies corresponding to
naive executions of PyTorch and TensorFlow (referred to as PT and TF
respectively) where the micro-batch size is equal to the mini-batch
size.

%%%%%%%%%%%%%%%%%%%%%%%%%%%%%%%%%%%%%%%%%

\noindent\underline{\Sys's MHA Implementation:}
As in \Sys's vgemm implementation on the Intel CPU backend, we offload
the computation of the dense inner tiles of the Proj1 and Proj2
operators in \Sys's MHA implementation on the ARM backends to gemm
calls in the OpenBLAS~\cite{openblas} library. Due to limitations of
our prototype implementation, however, offloading the computation this
way means that we cannot fuse the padding change operators with other
computational operators in this case. We see in
Fig.~\ref{fig:ap_per_ops_plot_arm}, however, that these pad fusion
operators are relatively cheap to perform on the CPU backend.

\begin{figure}
  \centering
  \includegraphics[width=0.95\columnwidth]{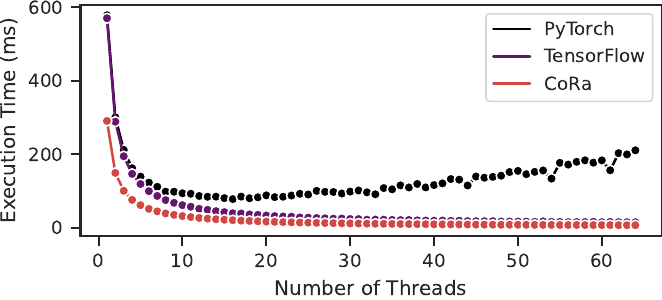}
  \caption{Execution latencies of PT, TF and \Sys~as the number of
  threads is increased for the MNLI dataset at a batch size of
  64. These measurements were performed on the 64-core CPU by changing
  the number of threads launched by OpenMP. Due to this, the
  measurements may not exactly be equal to the ones in
  Table~\ref{table:arm_fuller_eval}.}
  \label{fig:arm_scalability}
\end{figure}

%%%%%%%%%%%%%%%%%%%%%%%%%%%%%%%%%%%%%%%%%
\noindent\underline{Overall Performance Comparison:}
Table~\ref{table:arm_fuller_eval} shows the inference latencies for
the PyTorch, TensorFlow and \Sys~implementations of the MHA module on
the 8- and 64-core ARM CPUs. We saw that TF-UB trades-off parallelism
for reduced wasted computation. It, therefore, performs the best when
there is high parallelism in the workload (i.e. for datasets with
longer sequence lengths at higher batch sizes) and it performs the
worst when the workload has low parallelism (i.e. for datasets with
shorter sequences at lower batch sizes). This is because in the
presence of high parallelism in the workload, TF-UB can reduce the
micro-batch size much more (leading to much lower wasted padding) as
compared to the case of a workload with low parallelism. This is seen
reflected in the optimal micro-batch sizes shown in
Table~\ref{table:arm_fuller_eval}. TF-UB also performs better on the
8-core CPU which exposes lower parallelism as compared to the 64-core
CPUs. This is again reflected in the optimal micro-batch sizes which
are generally higher (leading to higher padding) on the 64-core CPU as
compared the 8-core CPU. Overall, we see that TF-UB and \Sys~perform
similarly on the 8-core ARM CPU, while \Sys~outperforms TF-UB by about
1.37$\times$ as the hardware parallelism increases on the 64-core
CPU. In both the cases, \Sys~performs significantly better than the TF
configuration of executing TensorFlow.

On the 8-core CPU, PyTorch in the PT-UB configuration performs better
than both TF-UB and \Sys~for datasets with higher sequence
lengths. Similar to TF-UB, PT-UB can more effectively trade-off batch
parallelism in these cases due to the high parallelism. Overall,
across all the datasets and batch sizes evaluated, \Sys~and PT-UB
perform similarly, while TF-UB is about 6\% slower than both on the
8-core CPU. We find that on the 64-core CPU, however, PyTorch's
performance does not scale well with the number of cores (this is
apparent in Fig.~\ref{fig:arm_scalability}) as compared to TensorFlow
and \Sys. Therefore, below, we only consider TensorFlow for further
analysis.
%%%%%%%%%%%%%%%%%%%%%%%%%%%%%%%%%%%%%%%%%

\noindent\underline{Per-Operator Execution Time Breakdown:}
Let us now look more closely at the execution times of the TensorFlow
and \Sys~implementations. Fig.~\ref{fig:ap_per_ops_plot_arm} provides
a breakdown of the execution times for four cases: (1) the MNLI
dataset at a batch size of 128 and the Wiki128 dataset at a batch size
of 32, which have the most and the least potential for savings on
wasted computation due to padding as Fig.~\ref{fig:flop_ratios} shows,
and (2) the RACE dataset at a batch size of 128 and the CoLA dataset
at a batch size of 32, which represent the best and worst cases for
the TF-UB configuration.

TF-UB and TF perform similarly for the CoLA dataset at batch size 32,
as that represents the worst case for TF-UB, and on the Wiki128
dataset at batch size 128 as there is little potential for
computational savings due to reduction in padding for that case. In
the remaining two cases, TF-UB performs better than TF as
expected. For the RACE dataset at batch size 128, which represents the
best case for TF-UB, TF-UB performs slightly better than \Sys. In
cases where \Sys~performs better than TensorFlow, we find that a lot
of the reduction in \Sys's absolute execution time stems from
computational savings in the Proj1 and Proj2 two operators, which
consume a significant portion of the execution time. The
QK\textsuperscript{T} and AttnV operators, however, show a higher
relative reduction in execution time as they are quadratically
proportional to the sequence lengths as opposed to Proj1 and Proj2
which are linearly proportional to sequence lengths. This difference
in proportionality is also reflected in the data for the Wiki128
dataset. TensorFlow generally does well on the Softmax operator,
performing better than \Sys~for the RACE and Wiki128 datasets. We
believe this is due to better optimized implementations and that this
gap can be reduced with more time spent optimizing \Sys's
implementation of the operator.

%% \newcommand{\PreserveBackslashTable}[1]{\let\temp=\\#1\let\\=\temp}
%% \newcolumntype{C}[1]{>{\PreserveBackslashTable\centering}m{#1}}
\newcommand{\ftnqkvmm}{5.4}
\newcommand{\ftpqkvmm}{7.16}
\newcommand{\coraqkvmm}{6.2}
\newcommand{\ftnqkvbp}{1.21}
\newcommand{\ftpqkvbp}{1.39}
\newcommand{\ftnqkt}{2.64}
\newcommand{\ftpqkt}{2.65}
\newcommand{\coraqkt}{2.12}
\newcommand{\ftnsoft}{4.08}
\newcommand{\ftpsoft}{4.08}
\newcommand{\corasoft}{1.93}
\newcommand{\ftnattnv}{2.79}
\newcommand{\ftpattnv}{2.78}
\newcommand{\coraattnv}{2.44}
\newcommand{\ftntrremp}{0.29}
\newcommand{\ftptrremp}{0.78}
\newcommand{\ftnpstl}{1.82}
\newcommand{\ftppstl}{2.42}
\newcommand{\corapstl}{2.31}
\newcommand{\ftnpstlbp}{0.38}
\newcommand{\ftppstlbp}{0.52}
\newcommand{\corapstlbp}{0.31}
\newcommand{\ftnffomm}{6.92}
\newcommand{\ftpffomm}{9.52}
\newcommand{\coraffomm}{8.06}
\newcommand{\ftnffba}{0.98}
\newcommand{\ftpffba}{1.38}
\newcommand{\ftnfftmm}{7.1}
\newcommand{\ftpfftmm}{9.47}
\newcommand{\corafftmm}{8.33}
\newcommand{\ftnfftbln}{0.38}
\newcommand{\ftpfftbln}{0.53}
\newcommand{\corafftbln}{0.31}
\newcommand{\ftnsum}{34.12}
\newcommand{\ftpsum}{42.82}
\newcommand{\corasum}{31.99}

\begin{table*}[htpb!]
  %% \vspace{-4.5cm}
  \centering \scriptsize \caption{Breakdown of the
  encoder layer execution time for FasterTransformer and \Sys~on the
  Nvidia GPU backend for the RACE dataset at batch size 128. Per-layer
  prelude code overheads are included in these latencies
  for \Sys. Both FasterTransformer and \Sys~implementations normally
  execute CUDA kernels asynchronously. For the purposes of profiling
  (i.e., this table only), these calls were made synchronous, which
  can lead to slower execution. We also show the end-to-end execution
  times under profiling for reference.}
  %% \begin{tabular}{C{2.3cm} C{2.3cm} C{0.9cm} C{1.1cm} C{0.9cm} C{2.3cm}}
  \begin{tabular}{c|ccccc}
    \toprule
    Op sub-graphs                               & FT~Ops     &  FT & FT-Eff & \Sys & \Sys~Ops \\ \midrule
    \multirow{2}{2.3cm}{\centering Proj1}       & QKV Proj. MM                              & \ftpqkvmm & \ftnqkvmm & \multirow{2}{1.3cm}{\centering \coraqkvmm} & \multirow{2}{2.5cm}{\centering QKV Proj.} \\
                                                & QKV Bias + AddPad                         & \ftpqkvbp & \ftnqkvbp & &  \\ \midrule
    QK\textsuperscript{T}                       & QK\textsuperscript{T}                     & \ftpqkt & \ftnqkt & \coraqkt & AddPad + QK\textsuperscript{T} \\ \midrule
    Softmax                                     & Softmax  & \ftpsoft & \ftnsoft & \corasoft & ChangePad + Softmax + ChangePad \\ \midrule
    AttnV                                       & AttnV                                     & \ftpattnv & \ftnattnv & \coraattnv & AttnV \\ \midrule
    \multirow{3}{2.3cm}{\centering Proj2}       & Transpose + RemovePad                     & \ftptrremp & \ftntrremp & & \\
                                                & Linear Proj. MM                           & \ftppstl & \ftnpstl & \corapstl & RemovePad + Linear Proj. MM + Bias + ResidualAdd \\
                                                & Linear Proj. Bias + ResidualAdd + LayerNorm     & \ftppstlbp & \ftnpstlbp & \corapstlbp & LayerNorm \\ \midrule
    \multirow{2}{2.3cm}{\centering FF1}         & FF1 MM                                    & \ftpffomm & \ftnffomm & \multirow{2}{1.3cm}{\centering \coraffomm} & \multirow{2}{2.5cm}{\centering FF1 MM + Bias + Activation} \\
                                                & FF1 Bias + Activation                     & \ftpffba & \ftnffba &  &  \\ \midrule
    \multirow{2}{2.3cm}{\centering FF2}         & FF2 MM                                    & \ftpfftmm & \ftnfftmm & \corafftmm & FF2 MM + Bias + ResidualAdd \\
                                                & FF2 Bias + ResidualAdd + LayerNorm        & \ftpfftbln & \ftnfftbln & \corafftbln & LayerNorm \\ \midrule
                                                & Total Execution Time                      & \ftpsum & \ftnsum & \corasum & Total Execution Time \\ \bottomrule

  \end{tabular}
  \label{table:ap_op_times}
\end{table*}

\clearpage
\appendix
\lstset{
  columns=flexible,
  basicstyle=\small\ttfamily,
  mathescape=true,
  escapeinside=||
}

\section{Artifact Appendix}

%%%%%%%%%%%%%%%%%%%%%%%%%%%%%%%%%%%%%%%%%%%%%%%%%%%%%%%%%%%%%%%%%%%%%
\subsection{Abstract}
This appendix describes how to reproduce the results described above
in \S\ref{sec:eval} and \S\ref{sec:ap_eval}. The experiments in the
paper evaluate \Sys~on an Nvidia V100 GPU, an 8-core, 16-thread Intel
CascadeLake CPU, an 8-core ARM Neoverse N1 CPU and a 64-core ARM
Neoverse N1 CPU. Below, we provide instructions to set up and execute
\Sys~as well as other frameworks used in the evaluation on each of
these platforms. As we start with publicly available Docker containers
in all cases, only a few GPU-related dependencies (such as cuDNN) as
well as \Sys's dependencies (such as the Z3 SMT solver, LLVM and
OpenBLAS) need to be installed. We provide instructions for each of
these. In each case, 100 GB of disk space should be more than enough
for the evaluation.

\subsection{Artifact check-list (meta-information)}
{
\begin{itemize}
  \item {\bf Compilation:} We reply on the publicly available
    compilers \verb+nvcc+ (the CUDA compiler to compile CUDA code
    generated by \Sys), \verb+gcc+ (to build \Sys~and other
    dependencies) and LLVM (to facilitate \Sys's code generation for
    CPUs).
  \item {\bf Data set:} We use sequence lengths for 8 different
    commonly used NLP datasets, all of which are included in the
    repositories described below.
  \item {\bf Run-time environment:} The artifact has been tested on
    Ubuntu 20.04 and with the following versions of different
    dependencies.
  \item {\bf Hardware:} We use an Nvidia V100 GPU, an 8-core,
    16-thread Intel CascadeLake CPU, an 8-core ARM Neoverse N1 CPU and
    a 64-core ARM Neoverse N1 CPU for our evaluation.
  \item {\bf Metrics:} We use execution time as the primary execution
    metric of evaluation.
  \item {\bf Output:} We generate CSV files, and also provide Python
    scripts to generate plots from this raw data.
  \item {\bf Experiments:} Python/bash scripts are provided to
    replicate the results.
  \item {\bf Disk space required:} 100 GB on each backend.
  \item {\bf Time needed to prepare workflow:} A few hours on each
    backend.
  \item {\bf Time needed to complete experiments:} Running all
    experiments on a backend takes several hours. The experiments on
    the ARM CPUs are particularly slow, taking 2-3 days to complete.
  \item {\bf Public availability:} Yes, in the form of GitHub
    repositories (linked later) as well as on Zenodo
    (\url{https://doi.org/10.5281/zenodo.6326455}).
\end{itemize}

%%%%%%%%%%%%%%%%%%%%%%%%%%%%%%%%%%%%%%%%%%%%%%%%%%%%%%%%%%%%%%%%%%%%%
\subsection{Description}

\subsubsection{How delivered}
Source code in the form of Github repositories and archived on Zenodo.

\subsubsection{Hardware dependencies}
We use an Nvidia V100 GPU, an 8-core, 16-thread Intel CascadeLake CPU,
an 8-core ARM Neoverse N1 CPU and a 64-core ARM Neoverse N1 CPU for
our evaluation.

\subsubsection{Software dependencies}
\label{sec:environment}
\noindent \textbf{GPU:} Below, we describe the environment we use
across the different hardware backends we evaluate \Sys~on. On all of
the backends, we use Ubuntu 20.04. Some of the frameworks below are
already installed as part of the Docker images we start with
(described below), while some need to be manually or installed. This
is described below in \S\ref{sec:installation}.

\noindent\textbf{Dependencies Common across Backends:} \Sys~requires
the following frameworks on all platforms: Z3 4.8.8, LLVM 9.0.0, cmake
$\ge$ 3.5 and g++ $\ge$ 5.0.

\noindent\textbf{Nvidia GPU:} CUDA 11.1 (V11.1.105), cuDNN 8.2.1,
PyTorch 1.9.0+cu111, FasterTransformer (modified on top of
FasterTransformer v4.0 (commit \verb+dd4c071+) and provided as part of
the \path{cora_benchmarks} repository). Make sure that \texttt{nvcc}
is on the \verb+PATH+.

\noindent\textbf{ARM CPUs:} OpenBLAS 0.3.10, PyTorch 1.10.0 with ARM
Compute Library 21.12, TensorFlow 2.6.0 with ARM Compute Library
21.09.

\noindent\textbf{Intel CPU:} Intel oneAPI MKL v2021.3.

\subsubsection{Data sets}
All datasets are included in the \verb+cora_benchmarks+ repository.

%%%%%%%%%%%%%%%%%%%%%%%%%%%%%%%%%%%%%%%%%%%%%%%%%%%%%%%%%%%%%%%%%%%%%
\subsection{Installation}
\label{sec:installation}
Refer to the file \path{ae_appendix_supplement.pdf} at the root of the
\verb+cora+ repository.

%%%%%%%%%%%%%%%%%%%%%%%%%%%%%%%%%%%%%%%%%%%%%%%%%%%%%%%%%%%%%%%%%%%%%
\subsection{Experiment workflow}
As we saw above, \Sys~is a tensor compiler. It takes as input a
description of a tensor operator and generates LLVM or CUDA code for
the kernel depending on whether the target is a CPU or a GPU. For
evaluating individual kernels, such as trmm, vgemm or any of the nine
kernels that make up the transformer layer (illustrated in
Figure~\ref{fig:fusion_graph} of the paper), merely executing the
python script implementing the kernel will compile the kernel,
generate the code and execute it. For evaluating the entire
transformer layer, or parts of it (such the self-attention or the MHA
modules), we separate the steps of code generation and execution
(\S\ref{sec:ap_impl}). We first generate compiled code for each of the
operators in the form of shared libraries, which are then loaded and
executed to form the layer to benchmark layer performance. The scripts
provided automate all of these steps.

%%%%%%%%%%%%%%%%%%%%%%%%%%%%%%%%%%%%%%%%%%%%%%%%%%%%%%%%%%%%%%%%%%%%%
\subsection{Evaluation Notes and Instructions}
Refer to the file \path{ae_appendix_supplement.pdf} at the root of the
\verb+cora+ repository for instructions on how to perform the
evaluation on the platforms referred to in the paper.

%%%%%%%%%%%%%%%%%%%%%%%%%%%%%%%%%%%%%%%%%%%%%%%%%%%%%%%%%%%%%%%%%%%%%
\subsection{Experiment customization}
\begin{enumerate}
  \item \textbf{GPU Evaluation:} The GPU evaluation has been tested on
    Nvidia GPUs with compute capability 70.  While the evaluation may
    run on other Nvidia GPUs as well, we have not yet tested it
    thoroughly. \Sys~can currently only support Nvidia GPUs as it only
    supports the generation of CUDA code.
  \item \textbf{CPU Evaluation:} The Intel and the ARM CPU evaluation
    should work for other CPUs beyond the CascadeLake and the Neoverse
    N1 CPUs we have tested it on. However, given that our schedules
    have been tuned for these CPUs, the performance might not be
    optimal. Further, when running on other CPUs, the LLVM target
    triple would need to be changed (in the file
    \path{cora_benchmarks/scripts/common.py} on line 10). The number
    of threads would also be need to be changed for PyTorch evaluation
    in the files
    \path{cora_benchmarks/bert_layer/pytorch/layer_cpu_micro_batch.py}
    and \path{cora_benchmarks/bert_layer/pytorch/layer_cpu.py} on the
    appropriate branch of the \path{cora_benchmarks} repository.
\end{enumerate}

%%%%%%%%%%%%%%%%%%%%%%%%%%%%%%%%%%%%%%%%%%%%%%%%%%%%%%%%%%%%%%%%%%%%%%%%%%%%%%%
%%%%%%%%%%%%%%%%%%%%%%%%%%%%%%%%%%%%%%%%%%%%%%%%%%%%%%%%%%%%%%%%%%%%%%%%%%%%%%%

\end{document}